\documentclass{article}

\usepackage[dblblindworkshop, final]{neurips_2026}
\workshoptitle{Reinforcement Learning for Experimental Sciences (RL4XS)}

\usepackage[utf8]{inputenc} % allow utf-8 input
\usepackage[T1]{fontenc}    % use 8-bit T1 fonts
\usepackage{hyperref}       % hyperlinks
\usepackage{url}            % simple URL typesetting
\usepackage{booktabs}       % professional-quality tables
\usepackage{amsfonts}       % blackboard math symbols
\usepackage{nicefrac}       % compact symbols for 1/2, etc.
\usepackage{microtype}      % microtypography
\usepackage{xcolor}         % colors
\usepackage{amsmath}
\usepackage{graphicx}
\usepackage{algorithm}
\usepackage{algpseudocode}
\usepackage{amssymb}
\usepackage{mathtools}
\usepackage{mathrsfs}
\usepackage{subcaption}
\usepackage{url}
\usepackage[space]{grffile}
\usepackage{multirow}
\usepackage{tcolorbox}
\usepackage{setspace}
\usepackage[normalem]{ulem}
\usepackage{wrapfig}

\newtheorem{pseudo}{Pseudocode}

\title{Batch-to-Streaming Deep Reinforcement Learning for Continuous Control}

\author{%
  Riccardo De Monte\thanks{Equal contribution.} \\
  Department of Information Engineering\\
  University of Padova \\
  \texttt{riccardo.demonte@phd.unipd.it} \\
  \And
  Matteo Cederle\footnotemark[1] \\
  Department of Information Engineering\\
  University of Padova \\
  \texttt{matteo.cederle@phd.unipd.it} \\
  \AND
  Gian Antonio Susto \\
  Department of Information Engineering\\
  University of Padova \\
  \texttt{gianantonio.susto@unipd.it} \\
}

\begin{document}

\maketitle

\begin{abstract}

  State-of-the-art deep reinforcement learning (RL) methods have achieved remarkable performance in continuous control tasks, yet their computational complexity is often incompatible with the constraints of resource-limited hardware, due to their reliance on replay buffers, batch updates, and target networks. The emerging paradigm of streaming deep RL addresses this limitation through purely online updates. Existing streaming methods, however, are studied in isolation, trained from scratch until convergence, whereas on-device learning is most useful as a continuation of a pre-trained policy. In this work, we propose two novel streaming deep RL algorithms, Streaming Soft Actor-Critic (S2AC) and Streaming Deterministic Actor-Critic (SDAC), designed as streaming counterparts of SAC and TD3, with performance comparable to state-of-the-art streaming baselines on standard benchmarks. Building on them, we study the batch-to-streaming transition as a problem in its own right: we show that a naive transition can cause a severe and lasting loss of the pre-trained policy, trace the failure to a mismatch between the optimizers used on either side of the switch and show that aligning the optimizers
  enables streaming deep RL in settings such as Sim2Real. 
  \footnote{Code available at
  \url{https://github.com/mcederle99/batch2streamDRL.git}}
\end{abstract}

\section{Introduction}
\label{sec:intro}

Reinforcement learning (RL) has emerged as a powerful framework for sequential decision-making and optimal control. By adopting neural networks as powerful function approximators, deep reinforcement learning (DRL) extends the representational capacity of classical RL, solving complex control tasks that were previously intractable \citep{mnih2015human}. Recent developments in DRL have enabled strong performance even in highly complex scenarios, such as continuous state and action control, with promising applications in robotics and other high-dimensional dynamical systems \citep{schulman2017proximal, fujimoto2018addressing, haarnoja2018soft, fujimoto2023sale}.
In this context, replay buffers, leveraging either off-policy or on-policy data, are employed to mitigate temporal correlations in collected samples and to improve sample efficiency, while mini-batch updates contribute to more reliable and statistically stable updates. 

While batch updates and replay buffers are critical to the effectiveness of modern DRL approaches, their computational demands make them challenging to implement for on-board learning in resource-constrained systems, such as edge devices. Nevertheless, reinforcement learning holds significant potential for tiny robotics \citep{neuman2022tiny}, enabling adaptive, data-driven control in compact and resource-limited platforms that have already demonstrated promising applications in autonomous inspection \citep{de2018inverted} and search-and-rescue operations \citep{duisterhof2021sniffy}. Motivated by these challenges, recent works have explored streaming deep reinforcement learning, in which deep RL methods rely solely on online updates, precluding the use of replay buffers, batch updates, or the additional computational complexity introduced by one or more target networks \citep{vasan2024deep}, making them suitable for on-device training and efficient continual learning. In particular, while Action Value Gradient (AVG) \citep{vasan2024deep} requires careful per-environment tuning of hyperparameters such as learning rate and temperature $\alpha$ of the Maximum Entropy RL formulation \citep{ziebart2010modeling}, Elsayed et al.\ \cite{elsayed2024streaming} introduce the \textit{stream-x} family of algorithms for various RL tasks, among which Stream AC($\lambda$), designed for continuous action control, achieves strong performance on both the MuJoCo Gym \citep{todorov2012mujoco} and DM Control Suite \citep{tunyasuvunakool2020dm_control} benchmarks.

Despite these promising results, existing works evaluate streaming deep RL algorithms in isolation, training them from a random initialization until convergence. However, for many practical applications, training a deep RL agent entirely from scratch on-device is not a realistic goal \citep{wu2025simlauncher}: the large number of environment interactions required to learn a meaningful policy from random initialization \citep{elsayed2024streaming}, combined with the erratic exploratory behavior of an untrained agent, would subject the real hardware to considerable mechanical stress and risk of damage. Streaming deep RL is therefore most valuable not as a replacement for batch methods, but as a complement to them: a policy is first trained using a state-of-the-art batch algorithm and is then handed over to a streaming algorithm that keeps adapting it online, on-device, under a tight computational budget.

Two deployment scenarios make this pattern concrete. The first is adaptation under the \textit{Sim2Real} gap \citep{hofer2021sim2real}, where the policy must recover the performance lost to the mismatch between simulated and real dynamics. This is particularly relevant in practice, since the majority of current applications of RL to real-world robotics do not fully capture the constraints imposed by on-device deployment: they either apply the simulation-trained policy directly to the real system \citep{chen2023learning}, assume that on-device finetuning with batch methods is possible \citep{yin2025rapidly}, or rely on the not always realistic presence of a 
%remote
server performing all computation in real-time communication with the edge device \citep{wang2022real}.  The second scenario instead arises when, despite relatively capable on-board hardware, computational resources cannot be dedicated exclusively to the RL training process: a robotic system must simultaneously handle perception, planning, and control, each competing for the same budget. Under such conditions, it is necessary to fall back from batch RL, which offers stronger sample efficiency at a heavier computational cost, to streaming RL, which trades some statistical efficiency for a far lighter resource~demand. In both scenarios the decisive step is the same, and the literature has left it unexamined: the \emph{transition} from a batch-trained agent to a streaming learner. Given a policy pre-trained with TD3 \citep{fujimoto2018addressing} or SAC \citep{haarnoja2018soft}, the streaming learner must continue learning without losing the performance that pre-training secured.

A prerequisite for even attempting the transition is a streaming algorithm that is architecturally compatible with the batch method used for pre-training, and here existing methods fall short. Stream AC($\lambda$) performs strongly on standard robotics benchmarks, but it is an on-policy, state-value-based actor-critic, a design inherently misaligned with the state-of-the-art batch algorithms used in continuous control, such as TD3 \citep{fujimoto2018addressing} and SAC \citep{haarnoja2018soft}: a TD3 checkpoint provides an action-value critic that Stream AC($\lambda$) has no way to consume. Additionally, its compatibility with PPO \citep{schulman2017proximal} does not offer a viable alternative, as PPO is known to underperform in robotics and continuous control tasks \citep{fujimototowards}, a finding further corroborated by Elsayed et al.\ \cite[Figure~14]{elsayed2024streaming}.

In light of this, we propose two novel streaming deep RL algorithms, namely Streaming Deterministic Actor-Critic (SDAC) and Streaming Soft Actor-Critic (S2AC), designed to adapt TD3 and SAC for the streaming setting.
%designed as streaming counterparts of SAC and TD3.
Across twelve continuous-control benchmarks, both are competitive with Stream AC($\lambda$), with no method dominating. We further show that a naive batch-to-streaming transition can collapse the pre-trained policy, and trace the failure to an optimizer mismatch across the switch: TD3 and SAC use Adam \citep{kingma2014adam}, whereas streaming methods use Overshooting-bounded Gradient Descent (ObGD) \citep{elsayed2024streaming}. We prove that ObGD coincides with Stochastic Gradient Descent with Clipping (SGDC) \citep{zhang2020adaptive, sun2025revisiting} in the streaming regime, which points to a remedy requiring no change to the streaming algorithm: pre-train the critic with SGDC so that pre-training and finetuning share an optimizer family. Combined with a short critic warm-up, this eliminates the failure regime and lets on-device adaptation build on the pre-trained policy.

\section{Preliminaries}
\label{sec:prel}

We assume that the RL agent interacts with an environment in discrete timesteps. We model the overall problem as a fully-observable Markov decision process (MDP), defined by the tuple $(\mathcal{S},\mathcal{A},R,p,r,d_0, \gamma)$: $\mathcal{S}$ denotes the set of all possible states, $\mathcal{A}=\mathbb{R}^n$ denotes the continuous action space, $p:\mathcal{S}\times \mathcal{A}\to \Delta(\mathcal{S})$ denotes the unknown transition kernel, giving the distribution of the next state $\boldsymbol{s}_{t+1}\in\mathcal{S}$ conditioned on the current state $\boldsymbol{s}_t\in\mathcal{S}$ and action $\boldsymbol{a}_t\in\mathcal{A}$, and $d_0$ denotes the initial state distribution $\boldsymbol{s}_0\sim d_0$. At each time step $t$ the agent samples an action $\boldsymbol{a}_t\sim \pi(\cdot|\boldsymbol{s}_t)$ according to a policy $\pi:\mathcal{S}\to\Delta(\mathcal{A})$, in general stochastic.
Given $(\boldsymbol{s}_t,\boldsymbol{a}_t, \boldsymbol{s}_{t+1})$, the environment emits a stochastic reward $r_{t+1}\in R$ according to the distribution given by $r:\mathcal{S}\times \mathcal{A}\times \mathcal{S}\to \Delta(R)$. By denoting with $\gamma\in[0,1)$ the discount factor, the return at time $t$ is defined as the sum of discounted future rewards, namely, $G_t=\sum_{k=0}^{+\infty}\gamma^{k}r_{t+k+1}$.  %The value-function for a given policy $\pi$ and state $\boldsymbol{s}$ is defined as $V_\pi(\boldsymbol{s})=\mathbb{
%E}_\pi[G_t|\boldsymbol{s}_t=\boldsymbol{s}]$, while the action-value function for a pair $(\boldsymbol{s},\boldsymbol{a})\in \mathcal{S}\times \mathcal{A}$ is defined as $Q_\pi(\boldsymbol{s},\boldsymbol{a})=\mathbb{E}_\pi[G_t|\boldsymbol{s}_t=\boldsymbol{s},\boldsymbol{a}_t=\boldsymbol{a}]$. 
The value-function for a given policy $\pi$ and state $\boldsymbol{s}$ is defined as $V_\pi(\boldsymbol{s})=\mathbb{
E}_\pi[G_t|\boldsymbol{s}_t=\boldsymbol{s}]$, while the action-value function for a pair $(\boldsymbol{s},\boldsymbol{a})\in \mathcal{S}\times \mathcal{A}$ is defined as $Q_\pi(\boldsymbol{s},\boldsymbol{a})=\mathbb{E}_\pi[G_t|\boldsymbol{s}_t=\boldsymbol{s},\boldsymbol{a}_t=\boldsymbol{a}]$. The ultimate goal of any RL algorithm is to maximize the objective function $J(\pi)=\mathbb{E}_{\boldsymbol{s}_0\sim d_0}[V_{\pi}(\boldsymbol{s}_0)]$.
We define the soft value function as $V_{\pi}^{\text{soft}}(\boldsymbol{s})=\mathbb{E}_{\pi}[\sum_{k=0}^{+\infty}\gamma^k(r_{t+k+1}+\alpha\mathcal{H}(\pi(\cdot|\boldsymbol{s}_{t+k})))|\boldsymbol{s}_t=\boldsymbol{s}]$, and the soft action-value function $Q^{\text{soft}}_{\pi}(\boldsymbol{s},\boldsymbol{a})=\mathbb{E}_{\pi}[r_{t+1}+\gamma V_{\pi}^{\text{soft}}(\boldsymbol{s}_{t+1})|\boldsymbol{s}_t=\boldsymbol{s},\boldsymbol{a}_t=\boldsymbol{a}]$, where $\alpha\in(0,+\infty)$ is a temperature hyper-parameter that balances reward maximization against entropy maximization, controlling the sensitivity of the policy to differences in action-values. When the state space $\mathcal{S}$ becomes extremely large, a natural solution is to rely on powerful universal function approximators such as neural networks, and several works followed this idea leading to the Deep Reinforcement Learning research field. In this context, we denote by $\pi_{\boldsymbol{\theta}}$, with $\boldsymbol{\theta}\in\mathbb{R}^d$, a policy parametrized by a neural network. Additionally, many existing DRL approaches follow the Actor-Critic paradigm \citep{fujimoto2018addressing, schulman2017proximal, haarnoja2018soft}, where a critic network is used to approximate either the value function or the action-value function  to guide the actor updates (e.g. $V_{\pi_{\boldsymbol{\theta}}}(\cdot\,;\,\boldsymbol{\phi})\approx V_{\pi_{\boldsymbol{\theta}}}(\cdot)$ with $\boldsymbol{\phi}\in\mathbb{R}^m$).

\paragraph{Batch and Streaming DRL}
\label{subsec:streadrl} We refer to batch methods as model-free DRL approaches, such as PPO 
\citep{schulman2017proximal}, DDPG \citep{lillicrap2020continuous}, TD3 \citep{fujimoto2018addressing} and SAC \citep{haarnoja2018soft}, 
that perform policy and critic updates using a batch of samples 
$\mathcal{B}=\{(\boldsymbol{s}_t,\boldsymbol{a}_t,\boldsymbol{s}_{t+1},r_{t+1})\}_{i}$ 
drawn from a replay buffer $\mathcal{D}$, with $|\mathcal{D}|\ge |\mathcal{B}|$. 
In contrast, in streaming DRL \citep{vasan2024deep, elsayed2024streaming} learning 
is purely incremental: only the most recent transition 
$(\boldsymbol{s}_t,\boldsymbol{a}_t,\boldsymbol{s}_{t+1},r_{t+1})$ is used to 
update the neural networks, i.e., $|\mathcal{D}|=|\mathcal{B}|=1$, requiring 
minimal memory usage. Furthermore, while batch methods typically rely on target 
networks, introducing additional computational and memory overhead, streaming DRL 
foregoes their use entirely. In the context of continuous control, Vasan et al.\ 
\citep{vasan2024deep} propose Action Value Gradient (AVG), an on-policy streaming 
counterpart of SAC \citep{haarnoja2018soft}, while Elsayed et al.\ 
\citep{elsayed2024streaming} propose Stream AC$(\lambda)$, a streaming actor-critic 
algorithm that, unlike AVG, requires no per-environment hyperparameter tuning. 
Given these properties and its relevance to our approach, we introduce 
Stream AC$(\lambda)$ in detail in the following paragraph.

\paragraph{Stream AC($\lambda$)}\label{subsec:streamac}Elsayed et al.\ \citep{elsayed2024streaming} propose Stream AC$(\lambda)$, an actor-critic (AC) approach, where a critic network with parameters $\boldsymbol{\phi}\in\mathbb{R}^m$ is used to approximate the value function $V_{\pi_{\boldsymbol{\theta}}}(\boldsymbol{s};\boldsymbol{\phi})\approx V_{\pi_{\boldsymbol{\theta}}}(\boldsymbol{s})$. Concerning the critic training, rather than minimizing the squared of the one-step Temporal Difference (TD) error $\delta_t=r_{t+1}+\gamma\cdot \text{stop-grad}(V_{\pi_{\boldsymbol{\theta}}}(\boldsymbol{s}_{t+1};\boldsymbol{\phi}))-V_{\pi_{\boldsymbol{\theta}}}(\boldsymbol{s}_t;\boldsymbol{\phi})$, commonly used to perform an on-line update of the critic network, Stream AC$(\lambda)$ relies on TD$(\lambda)$ \citep{sutton1981toward}: defined the eligibility traces $\boldsymbol{z}_t=\gamma\lambda \boldsymbol{z}_{t-1}+\nabla_{\boldsymbol{\phi}}V_{\pi_{\boldsymbol{\theta}}}(\boldsymbol{s}_t;\boldsymbol{\phi})$, with $\boldsymbol{z}_{-1}=\boldsymbol{0}$ and $\lambda\in(0,1)$, the critic is updated as $\boldsymbol{\phi}\gets\boldsymbol{\phi}+\eta_V \delta_t\boldsymbol{z}_t$, where $\eta_V$ is the learning rate. The use of eligibility traces leads to better credit assignment while retaining the streaming nature of the update. The same philosophy is applied to the actor: whereas standard one-step Actor-Critic uses the estimator $\delta_t\nabla_{\boldsymbol{\theta}}\log\pi_{\boldsymbol{\theta}}(\boldsymbol{a}_t|\boldsymbol{s}_t)\approx \nabla_{\boldsymbol{\theta}}J(\boldsymbol{\theta})$ \citep{sutton1999policy}, Stream AC$(\lambda)$  maintains actor traces $\boldsymbol{e}_t=\lambda\gamma\boldsymbol{e}_{t-1}+\nabla_{\boldsymbol{\theta}}\log\pi_{\boldsymbol{\theta}}(\boldsymbol{a}_t|\boldsymbol{s}_t)$, and updates the policy 
%parameters 
as $\boldsymbol{\theta}\gets\boldsymbol{\theta}+\eta_\pi\delta_t\boldsymbol{e}_t$.
%Additionally,
Stream AC$(\lambda)$ also incorporates entropy regularization with eligibility traces.

A critical challenge in streaming actor-critic methods is training stability: unlike batch-based algorithms, which smooth gradient noise over large mini-batches, streaming updates operate on single transitions, making the optimization landscape considerably noisier. Additionally, Lyle et al.\ \cite{lyle2023understanding} have shown that Adam \citep{kingma2014adam} can be a source of instability in non-stationary settings. %, as stale second-moment estimates cause the effective learning rate to grow unboundedly, leading to overshooting.
This motivates Elsayed et al.\  \cite{elsayed2024streaming} to introduce Overshooting-bounded Gradient Descent (ObGD, Algorithm \ref{alg:ObGD_main}), an optimizer designed to avoid overshooting without adding extra computation as done by backtracking. It is worth noting that ObGD, originally proposed as optimizer for the critic network $V_{\pi_{\boldsymbol{\theta}}}(\cdot\,;\,\boldsymbol{\phi})$, can be applied to any regression problem. %Moreover, while ObGD is derived to avoid overshooting, it resembles SGD with Clipping (SGDC) \citep{zhang2020adaptive, sun2025revisiting}, a variant of SGD designed to ensure the convergence of SGD in heavy-tailed noise. We derive such connection in \autoref{sec:supp_sgdc_obgd}.
Notably, despite ObGD being derived to avoid overshooting, it closely resembles Stochastic Gradient Descent with Clipping (SGDC) \citep{zhang2020adaptive, sun2025revisiting}, a connection we formalize in \autoref{subsec:finetuneexp}.
% , a variant of SGD designed to ensure convergence under heavy-tailed noise, a connection we formalize in \autoref{subsec:finetuneexp}. \\
%\autoref{sec:supp_sgdc_obgd}.

\begin{algorithm}
\setstretch{1.2}
\caption{{\bf O}vershooting-{\bf b}ounded {\bf G}radient {\bf D}escent (ObGD) (Elsayed et al.\ \cite{elsayed2024streaming})}
\label{alg:ObGD_main}
\begin{algorithmic}[1]

\State \textbf{Given} Traces/Output gradient $\boldsymbol{z}$, error $\delta$, step size $\eta$, scaling factor $\kappa$ 

\State $\bar{\delta}=\max(|\delta|,1)$

\State $M\leftarrow \kappa 
\bar{\delta}||\boldsymbol{z}||_1$

\State $\eta\leftarrow\eta \cdot \min(1,\frac{1}{M})$

\State $\boldsymbol{\phi}\leftarrow\boldsymbol{\phi}+\eta \delta \boldsymbol{z}$

\end{algorithmic}
\end{algorithm}

Following common practice in supervised learning, normalizing the input data has been shown to improve training stability also in RL \citep{andrychowicz2020matters, engstrom2020implementation}. To this end, Stream AC($\lambda$) normalizes states following Andrychowicz et al.\ \citep{andrychowicz2020matters} and scales the reward signal as proposed by Engstrom et al.\ \citep{engstrom2020implementation}. %Specifically, the online algorithm of Welford \citep{welford1962note} is used to track the statistics required for state normalization and reward scaling. 
Moreover, Elsayed et al.\ \cite{elsayed2024streaming} employ sparse network initialization, and both the critic and actor networks incorporate LayerNorm \citep{ba2016layer}.

\section{Streaming algorithms compatible with batch RL}
\label{sec:method}

As argued in \autoref{sec:intro}, transferring a batch-trained agent to a streaming
learner presupposes a streaming algorithm that the batch checkpoint can be transferred
\emph{into}. This section introduces two such algorithms: Streaming Soft Actor-Critic
(S2AC) and Streaming Deterministic Actor-Critic (SDAC).
They are designed under the additional constraint of mirroring the objective and the network topology of a widely used batch algorithm, S2AC that of SAC
\citep{haarnoja2018soft}, SDAC that of TD3 \citep{fujimoto2018addressing}, so that the
actor and critic weights obtained during batch pre-training can be loaded directly into the streaming learner.

We first discuss the architectural and methodological choices shared by both methods, before turning to the algorithm-specific design decisions that were necessary to ensure stable and effective learning in practice.
As noted by Elsayed et al.\ \cite{elsayed2024streaming}, streaming deep reinforcement learning is primarily hindered by three challenges: instabilities arising from occasional large updates, activation nonstationarity, and improper data scaling. To address these issues, both S2AC and SDAC adopt sparse network initialization, LayerNorm \citep{ba2016layer} applied to the pre-activations of each layer, and the state normalization and reward scaling schemes of Andrychowicz et al.\ \cite{andrychowicz2020matters} and by Engstrom et al.\ \cite{engstrom2020implementation}, respectively. Pseudocodes for state normalization and reward scaling are reported in \autoref{sec:supp_norm}. While we omit any additional notation to highlight the state normalization, we will denote with $\sigma_r$ the time-varying reward scaling computed as by Engstrom et al.\ \cite{engstrom2020implementation}.

\paragraph{Streaming Soft Actor-Critic (S2AC)}
\label{subsec:s2ac}

We parameterize the actor as a stochastic policy $\pi_{\boldsymbol{\theta}}(\cdot|\boldsymbol{s}_t)$ 
and the critic as a neural network approximation $Q_{\pi_{\boldsymbol{\theta}}}^{\text{soft}}
(\boldsymbol{s}_t,\boldsymbol{a}_t;\boldsymbol{\phi})$ of the soft Q-function $Q_{\pi_{\boldsymbol{\theta}}}^{\text{soft}}
(\boldsymbol{s}_t,\boldsymbol{a}_t)$, where $\boldsymbol{\theta} \in \mathbb{R}^d$ and 
$\boldsymbol{\phi} \in \mathbb{R}^m$ denote the neural network weights. The critic is updated at each time step by minimizing the instantaneous squared soft Bellman residual:
\begin{align}
    \label{eq:softtderror}
    \ell_{Q^{\text{soft}}}(\boldsymbol{s}_t,\boldsymbol{a}_t;\boldsymbol{\phi}) = &\Biggr(Q_{\pi_{\boldsymbol{\theta}}}^{\text{soft}}(\boldsymbol{s}_t,\boldsymbol{a}_t;\boldsymbol{\phi}) - \Biggr(\cfrac{r(\boldsymbol{s}_t,\boldsymbol{a}_t)}{\sigma_r} + \\ \nonumber 
    &+\gamma\mathbb{E}_{\boldsymbol{s}_{t+1}\sim p,\, \boldsymbol{a}_{t+1}\sim\pi_{\boldsymbol{\theta}}}\Bigr[Q_{\pi_{\boldsymbol{\theta}}}^{\text{soft}}(\boldsymbol{s}_{t+1},\boldsymbol{a}_{t+1};\boldsymbol{\phi}) - \alpha\log\pi_{\boldsymbol{\theta}}(\boldsymbol{a}_{t+1}|\boldsymbol{s}_{t+1})\Bigr]\Biggr)\Biggr)^2,
\end{align}
where the target is computed using a single next transition $(\boldsymbol{s}_{t+1}, \boldsymbol{a}_{t+1})$, 
reducing the expectation to a one-sample estimate and making the update solely dependent on the tuple 
$(\boldsymbol{s}_t, \boldsymbol{a}_t, r_{t+1}, \boldsymbol{s}_{t+1})$, as in classical RL algorithms \citep{sutton1998introduction, watkins1992q, konda1999actor}.
%where the transition $(\boldsymbol{s}_t, \boldsymbol{a}_t)$ is collected online under the current policy, and $\sigma_r$ denotes the running standard deviation of the observed rewards, computed following~\cite{engstrom2020implementation}. 
In contrast to standard SAC, and consistent with the streaming nature of our algorithm, we forgo the use of target networks, relying instead on the online estimate of $Q_{\pi_{\boldsymbol{\theta}}}^{\text{soft}}(\cdot,\cdot;\boldsymbol{\phi})$ when computing the Bellman target. Critic updates are performed using eligibility traces in conjunction with the ObGD optimizer, modifying the original residual described in \autoref{subsec:streamac} with the soft Bellman residual in \autoref{eq:softtderror}.

For the actor, we consider the policy improvement objective proposed by 
\cite{haarnoja2018soft}:
\begin{equation}
    J_\pi(\boldsymbol{s}_t;\boldsymbol{\theta}) = \mathbb{E}_{\boldsymbol{a}_t\sim\pi_{\boldsymbol{\theta}}}\left[ \alpha \log\left(\pi_{\boldsymbol{\theta}}(\boldsymbol{a}_t|\boldsymbol{s}_t)\right) - Q_{\pi_{\boldsymbol{\theta}}}^{\text{soft}}(\boldsymbol{s}_t, \boldsymbol{a}_t;\boldsymbol{\phi}) \right],
    \label{eq:policy_objective}
\end{equation}
which trades off expected return maximization and entropy regularization 
through the temperature parameter $\alpha$. To enable efficient gradient-based optimization, we apply the reparameterization trick, expressing sampled actions as a deterministic transformation of noise: $\boldsymbol{a}_t = f(\boldsymbol{\epsilon}_t, \boldsymbol{s}_t;\boldsymbol{\theta})$, where $\boldsymbol{\epsilon}_t \sim \mathcal{N}(\boldsymbol{0}, \boldsymbol{I})$. This yields the tractable objective:
\begin{equation}
    J_\pi(\boldsymbol{s}_t;\boldsymbol{\theta}) = \mathbb{E}_{\boldsymbol{\epsilon}_t\sim\mathcal{N}} \left[ \alpha \log \pi_{\boldsymbol{\theta}}(f(\boldsymbol{\epsilon}_t,\boldsymbol{s}_t;\boldsymbol{\theta})|\boldsymbol{s}_t) - Q_{\pi_{\boldsymbol{\theta}}}^{\text{soft}}(\boldsymbol{s}_t, f(\boldsymbol{\epsilon}_t,\boldsymbol{s}_t;\boldsymbol{\theta});\boldsymbol{\phi}) \right],
    \label{eq:reparam_objective}
\end{equation}
where the expectation is approximated with a single action sample $\boldsymbol{a}_t = 
f(\boldsymbol{\epsilon}_t,\boldsymbol{s}_t;\boldsymbol{\theta})$, consistent with the 
streaming setting. Unlike the critic, the actor is updated without eligibility traces, employing the Adam optimizer~\citep{kingma2014adam}, similarly to batch deep RL algorithms.

Finally, we recall that the entropy regularization coefficient $\alpha$ acts as a temperature parameter, controlling the tradeoff between reward and entropy maximization: higher values encourage more stochastic, exploratory policies, while lower values steer the agent towards more deterministic, reward-focused behavior.
An important consequence of reward normalization arises when considering the role of $\alpha$ during training. Since rewards are scaled by the running standard deviation $\sigma_r$, the effective magnitude of the reward signal fluctuates over time. As a result, a fixed value of $\alpha$ can lead to undesirable behavior: when $\sigma_r \ll 1$, the normalized rewards are amplified, diminishing the relative contribution of the entropy term and pushing the policy towards excessive determinism; conversely, when $\sigma_r \gg 1$, the reward signal is suppressed, causing the entropy term to dominate and yielding an overly stochastic policy. In either case, the balance between the two objectives, which $\alpha$ is designed to maintain, is inadvertently disrupted by the normalization.
To address this issue, we propose scaling the entropy coefficient by the same factor used to normalize the rewards. Writing $\alpha_0$ for the base value, which is the quantity actually chosen per domain, the coefficient entering the objectives above becomes time-varying, $\alpha \leftarrow \alpha_0/\sigma_r$. This correction ensures that the relative weighting between the reward and entropy terms remains consistent throughout training, regardless of the current reward statistics. The practical impact of this design choice is empirically validated in the ablation study presented in \autoref{sec:supp_abl_stream}. For a pseudocode of S2AC instead, refer to \autoref{sec:supp_pseudocodes}.

\paragraph{Streaming Deterministic Actor-Critic (SDAC)}
\label{subsec:sdac}

Here instead, to guarantee compatibility with 
%DDPG and
TD3, we follow the
Q-learning approach by parameterizing the action-value function as a
neural network approximation $Q(\boldsymbol{s}_t,\boldsymbol{a}_t;\boldsymbol{\phi})$
of $Q(\boldsymbol{s}_t,\boldsymbol{a}_t)$, where $\boldsymbol{\phi}\in\mathbb{R}^m$
denotes the critic network parameters. Due to the continuous action space, we
follow Silver et. al.    \cite{silver2014deterministic} in defining a deterministic policy
$\pi_{\boldsymbol{\theta}}(\boldsymbol{s}_t)$, where $\boldsymbol{\theta}\in\mathbb{R}^d$
denotes the actor network parameters. To encourage exploration, deterministic
actions are augmented with Gaussian noise as
$\boldsymbol{a}_t=\pi_{\boldsymbol{\theta}}(\boldsymbol{s}_t)+\boldsymbol{\epsilon}_1$,
where $\boldsymbol{\epsilon}_1\sim\mathcal{N}(\boldsymbol{0},\sigma^2\boldsymbol{I})$.
SDAC is thus, to the best of our knowledge, the first off-policy streaming deep RL
algorithm built on a deterministic policy. Given the deterministic nature of the policy and the intractability of solving
$\arg\max_{\boldsymbol{a}} Q(\boldsymbol{s},\boldsymbol{a};\boldsymbol{\phi})$, the
policy network is optimized to approximately solve this optimization problem; in
particular, we employ the Deterministic Policy Gradient theorem (DPG,
\citealp{silver2014deterministic}) for the policy update:
\begin{equation}
    \nabla_{\boldsymbol{\theta}} J_\pi(\boldsymbol{s}_t;\boldsymbol{\theta}) =  \nabla_{\boldsymbol{a}_t} Q(\boldsymbol{s}_t, \boldsymbol{a}_t;\boldsymbol{\phi}) \big|_{\boldsymbol{a}_t=\pi_{\boldsymbol{\theta}}(\boldsymbol{s}_t)} \nabla_{\boldsymbol{\theta}} \pi_{\boldsymbol{\theta}}(\boldsymbol{s}_t).
\end{equation}
Concerning the critic updates, we employ Q-learning by minimizing the following
instantaneous squared residual
\begin{equation}
    \label{eq:tderror}
    \ell_{Q}(\boldsymbol{s}_t,\boldsymbol{a}_t;\boldsymbol{\phi}) = \Biggl(Q(\boldsymbol{s}_t,\boldsymbol{a}_t;\boldsymbol{\phi}) - \Biggl(\frac{r(\boldsymbol{s}_t,\boldsymbol{a}_t)}{\sigma_r} + \gamma\mathbb{E}_{\boldsymbol{s}_{t+1}\sim p,\, \boldsymbol{\epsilon}_2\sim\mathcal{N}}\Bigl[Q(\boldsymbol{s}_{t+1},\pi_{\boldsymbol{\theta}}(\boldsymbol{s}_{t+1})+\boldsymbol{\epsilon}_2;\boldsymbol{\phi})\Bigr]\Biggr)\Biggr)^2,
\end{equation}
where, inspired by Fujimoto et al. \cite{fujimoto2018addressing}, the target includes additional
noise $\boldsymbol{\epsilon}_2\sim\mathcal{N}(\boldsymbol{0},\sigma^2\boldsymbol{I})$
to mitigate the risk of the critic overfitting to narrow peaks in the value
estimate. The injection of small Gaussian noise into the target encourages the
Q-value function to be smooth in the neighborhood of each action, reducing the
variance introduced by function approximation errors and improving the stability
of the learning targets. The practical importance of this design choice is
empirically validated in the ablation study of \autoref{sec:supp_abl_stream}.
As for S2AC, we employ ObGD with traces fort the critic and no target network is used; that is, the
target in \autoref{eq:tderror} is given by the online critic
$Q(\cdot,\cdot;\boldsymbol{\phi})$ evaluated on a single next transition
$(\boldsymbol{s}_{t+1}, \boldsymbol{a}_{t+1})$, making the update depend solely on
the tuple $(\boldsymbol{s}_t, \boldsymbol{a}_t, r_{t+1}, \boldsymbol{s}_{t+1})$. For a pseudocode of SDAC, refer to \autoref{sec:supp_pseudocodes}, in the supplementary material.

\section{From batch to streaming}
\label{sec:b2s}

The previous section introduces streaming algorithms whose architecture, weight layout,
and objective match those of a batch pre-trained agent, so parameters
$(\boldsymbol{\theta}_0,\boldsymbol{\phi}_0)$ obtained during pre-training can be loaded directly into
a streaming learner that continues to update them from single transitions. We ask this
transition to be \emph{safe}, in that the return of the pre-trained policy is not lost at the
switch, and \emph{productive}, in that subsequent streaming updates improve on it, whether to
close a Sim2Real gap or to continue where batch training left off. Matching architecture
and objective is necessary but not sufficient for this: it enables the transfer but does not guarantee it. Below we identify the remaining sources of failure: observation
normalization and reward scaling, and the optimizer.

First, unlike TD3 and SAC, their streaming counterparts SDAC and S2AC apply observation
normalization and reward scaling; we remove this difference in \autoref{sec:batch_supp}
by augmenting the batch methods (\textit{TD3-norm} and \textit{SAC-norm}), tracking
statistics as in the streaming case while leaving buffer samples unprocessed and
normalizing each batch at every update. Second, the batch algorithm optimizes the critic
with Adam \citep{kingma2014adam} and the streaming learner with ObGD
\citep{elsayed2024streaming}; nothing forces the batch side to use Adam, and this is the
only difference concerning the \emph{history} of the pre-trained weights rather than the
mechanics of the update.

\paragraph{A naive transition fails}
\label{subsec:naive}

As a first attempt, one could be tempted to transition directly from TD3-norm to SDAC, retaining Adam as the critic optimizer during pre-training.
\begin{wrapfigure}{r}{0.37\textwidth}
  \centering
  \includegraphics[width=0.35\textwidth]{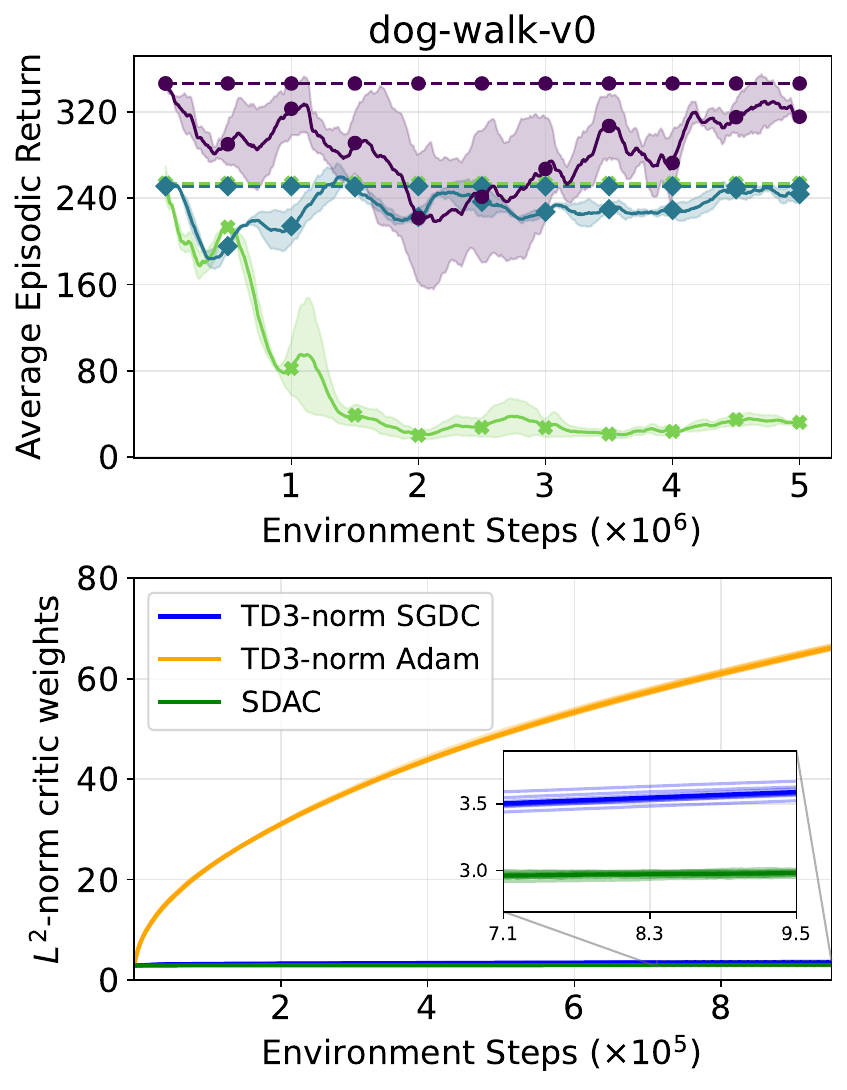}
  \caption{Top: finetuning performance of SDAC after pre-training with TD3-norm using Adam as critic optimizer. Bottom: $L^2$-norm of the critic's weights across training.}
  \label{fig:ft_adam_q_norm}
\end{wrapfigure}The top row of \autoref{fig:ft_adam_q_norm} reports the outcome: each of the three curves corresponds to SDAC finetuning started from a different TD3-norm pre-training checkpoint. As shown, this naive approach can lead to a severe and lasting drop in agent performance, unacceptable in a deployment setting.
Drawing the recent work of Pascanu et al. \cite{pascanu2025optimizers}, we hypothesize that the critic
optimizer shapes the learned solution through its inductive biases, and that a pre-trained
critic therefore carries properties of the optimizer that produced it. If those properties
are ill-suited to the optimizer that takes over at deployment, the transition can fail even
though the weights transfer without loss of information.
Some evidence of this is given by the bottom row of \autoref{fig:ft_adam_q_norm}, which shows the $L^2$-norm of the critic weights, growing by more than an order of magnitude under Adam
during pre-training, while remaining far smaller under ObGD in SDAC. Adam and ObGD therefore leave the
critic in very different regions of parameter space, and the streaming learner is built to operate in the latter. 

This suggests a remedy that leaves the streaming algorithm untouched and instead adjusts
pre-training, the side of the switch we still control: pre-training the critic with an
optimizer from the same family as the downstream one lets the streaming learner inherit a
critic it can continue to optimize. The next subsection shows such a family exists, and
that its batch member is a standard, well-studied optimizer.

\paragraph{ObGD and SGDC coincide in the streaming regime}
\label{subsec:obgd_sgdc}

We now formalize the connection between ObGD and Stochastic Gradient Descent with Clipping
(SGDC) \citep{zhang2020adaptive, sun2025revisiting} anticipated in
\autoref{subsec:streamac}. Despite being derived from unrelated motivations, ObGD to
bound overshooting in online updates, SGDC to guarantee convergence of SGD under
heavy-tailed gradient noise, the two coincide in the regime in which streaming
algorithms operate.

We briefly recall the notation of Algorithm~\ref{alg:ObGD_main}: when traces are used
($\lambda > 0$), $\boldsymbol{z}$ represents the eligibility traces vector; otherwise
($\lambda = 0$), it reduces to the gradient of the network output with respect to its
parameters. In particular, for the critic network of TD3 we have
$\boldsymbol{z}=\nabla_{\boldsymbol{\phi}}
Q(\boldsymbol{s},\boldsymbol{a};\boldsymbol{\phi})$). In general,
$\delta$ represents the regression error between a target $\hat{y}$ and the network output,
which in the case of TD3 takes the form
$\delta=\hat{y}-Q(\boldsymbol{s}_t,\boldsymbol{a}_t;\boldsymbol{\phi})$,
where $\hat{y}$ is the bootstrapped target (similarly for SAC where the soft action-value function is considered instead).

While ObGD is designed for regression tasks and online updates, SGDC
\citep{sun2025revisiting} is a general-purpose optimizer for any loss function and batch
size. Following Sun et al. \cite{sun2025revisiting}, given the gradient $\boldsymbol{g}$ of any loss
function over a batch of any size, the update rule is:
\begin{equation}
\boldsymbol{\phi}\leftarrow\boldsymbol{\phi}-\eta\cdot\text{Clip}_h(\boldsymbol{g}), \quad \text{Clip}_h(\boldsymbol{g})=\min\left(1,\frac{h}{\|\boldsymbol{g}\|}\right)\cdot \boldsymbol{g},
\end{equation}
where $\|\cdot\|$ is a generic norm, possibly the $L^1$-norm $\|\cdot\|_1$, and $h$ is a
hyperparameter. To establish the connection with ObGD, we specialize SGDC to the online
setting by considering a batch size of one and a squared loss
$\ell(\boldsymbol{\phi}) = \frac{1}{2}\delta^2$. Setting $h=1/2$ and adopting the
$L^1$-norm, the SGDC update becomes:
\begin{equation}\label{eq:sgdc-obg}
    \nabla_{\boldsymbol{\phi}}\frac{1}{2}\delta^2=-\delta \underbrace{\nabla_{\boldsymbol{\phi}}Q_{\pi_{\boldsymbol{\theta}}}(\boldsymbol{s}_t,\boldsymbol{a}_t;\boldsymbol{\phi})}_{\boldsymbol{z}}\;\implies\;\boldsymbol{\phi}\leftarrow\boldsymbol{\phi}+\eta\cdot \min\left(1,\frac{1}{2\cdot |\delta|\cdot \|\boldsymbol{z}\|_1}\right)\delta\boldsymbol{z}.
\end{equation}
Under the additional assumption $|\delta|>1$ and setting $\kappa=2$, as proposed by
Elsayed et al.\ \cite{elsayed2024streaming}, the update in \autoref{eq:sgdc-obg} reduces to that of ObGD,
showing that the two optimizers coincide in this regime.
Thus, SGDC is the natural choice for pre-training
a critic that will later be optimized with~ObGD.\\
The empirical confirmation of this analysis is shown in the bottom row of \autoref{fig:ft_adam_q_norm} and \autoref{fig:norm_walker}, which shows that the $L^2$-norm of the critic weights of TD3-norm under pre-training with SGDC remain at
a scale comparable to the one ObGD maintains within SDAC.

\section{Experiments}\label{sec:exp}

Following Elsayed et al.\ \cite{elsayed2024streaming}, we consider several environments from MuJoCo Gym
\citep{todorov2012mujoco} and DM Control Suite \citep{tunyasuvunakool2020dm_control} as
benchmarks for control in continuous action spaces. In \autoref{subsec:streamingexp} we
present the results achieved by training the policy for 20M steps with S2AC and SDAC, comparing them against Stream AC($\lambda$) as the primary streaming
baseline; we additionally compare against AVG \citep{vasan2024deep} in \autoref{sec:supp_avg}, though only on the subset of environments considered in the original work, as AVG requires extensive per-environment hyperparameter tuning and was not evaluated on all the environments that we consider here. In \autoref{subsec:finetuneexp}, we present the results achieved in the batch-to-streaming scenarios.\\
Concerning the experiments from scratch, all reported plots are generated by repeating each experiment with 10 different random seeds, where the shaded area represents a 95\% confidence interval. The agents are evaluated every 10,000 time steps by computing the average undiscounted return over 10 evaluation episodes. During evaluation, a deterministic policy is used: no exploration noise is added, and for stochastic policies the mean action is selected.

\subsection{S2AC and SDAC from scratch}
\label{subsec:streamingexp}

S2AC and SDAC are designed under a constraint that Stream AC($\lambda$) does not have to satisfy, namely that a SAC or TD3 checkpoint must be loadable into them. The question this section answers is what that constraint costs. Concerning the hyperparameters and the architectures used for all streaming RL algorithms, we refer to \autoref{sec:supp_stream_param}.

\begin{figure}[h]
    \begin{center}
        \includegraphics[width=\textwidth]{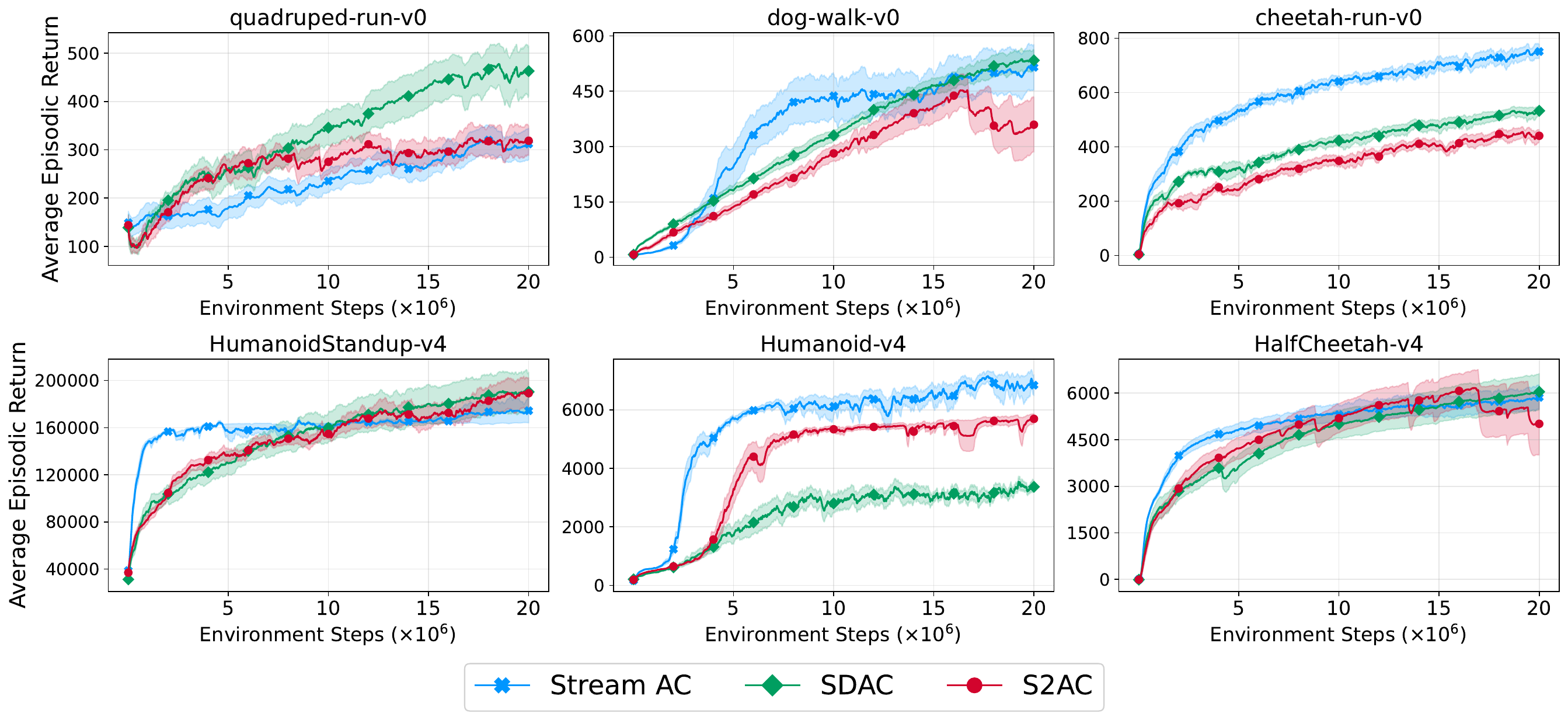}
    \end{center}
    \caption{Results for streaming DRL algorithms SDAC, S2AC, and Stream AC$(\lambda)$ on MuJoCo Gym and DM Control Suite tasks.}
    \label{fig:stream}
\end{figure}

As shown in \autoref{fig:stream}, and across the twelve environments obtained by including
the additional experiments of \autoref{sec:supp_stream_res}, S2AC and SDAC are competitive with Stream AC($\lambda$) without any method dominating, namely, the relative ordering is task-dependent. Regarding tuning, unlike AVG \citep{vasan2024deep}, neither method requires per-environment tuning of the optimizer, the learning rates, or the discount factor. SDAC introduces no environment-specific hyperparameters at all, while S2AC does require tuning of the base entropy coefficient, with the exact values reported in \autoref{table:s2ac_alpha}.

We report in \autoref{sec:supp_abl_stream} the ablation studies concerning the adaptive schedule $\alpha/\sigma_r$ and the target noise proposed in \autoref{sec:method}, the use of state normalization and reward scaling, the use of LayerNorm, and the use of ObGD as critic optimizer.

\subsection{Batch-to-streaming transfer}\label{subsec:finetuneexp}

\paragraph{Experimental setting}

The Sim2Real scenario discussed in \autoref{sec:intro} is emulated in simulation by perturbing
the dynamics of two DM Control Suite \citep{tunyasuvunakool2020dm_control} environments. We
derive one perturbed variant of \texttt{dog-walk-v0}, denoted \texttt{dog-walk-v0-p}, and two
of \texttt{walker-run-v0}, denoted \texttt{walker-run-v0-p1} and \texttt{walker-run-v0-p2}.
All modifications are listed in \autoref{sec:supp_modenv}.
% ; the corresponding \texttt{.xml}
% files are provided in the code repository.
In particular, the perturbations reduce joint stiffness and actuator authority, and are meant to
represent an ageing animal or the degradation of materials and actuators in a robotic platform.\\
% platform: reduced lumbar stiffness, for instance, corresponds to weakened back muscles in a
% biological dog or to fatigue in the spinal structure of a robotic one. The two walker
% variants differ in kind rather than degree: \texttt{-p1} perturbs both legs identically,
% whereas \texttt{-p2} degrades only the left leg, producing an asymmetric gait deficit that
% the pre-trained symmetric policy has no reason to anticipate.
% We emphasize that the Sim2Real scenario is thereby \emph{emulated} in simulation: a controlled change of dynamics isolates the effect we wish to study and keeps the comparison reproducible.
% , but it does not reproduce sensor noise, actuation delay, or the other sources of mismatch present on physical hardware. We return to this point in \autoref{sec:disc}.
In the second scenario, concerning batch-to-streaming alternation, a setting that arises naturally under energy or hardware temporal constraints, we consider environments that are particularly challenging for streaming approaches: \texttt{Humanoid-v4} and \texttt{HalfCheetah-v4} from MuJoCo Gym \citep{todorov2012mujoco}, and \texttt{quadruped-run-v0} from DM Control Suite \citep{tunyasuvunakool2020dm_control}, for which no perturbation is applied to the environment dynamics. 
%\paragraph{Pre-training with batch methods.}

As described in \autoref{sec:b2s}, to guarantee compatibility between a batch algorithm and its streaming counterpart, both TD3 and SAC are augmented with observation normalization and reward scaling. In \autoref{subsec:supp_batch_norm}, we report results for TD3-norm and SAC-norm, showing the effectiveness of normalization, as also previously shown by Lee et al. \cite{lee2024simba} for the SimBa architecture. We additionally report the effect of using SGDC as the critic optimizer in \autoref{subsec:supp_batch_mix}, where a reduction in sample efficiency is observed in a single environment. In both the \textit{Sim2Real} and batch-to-streaming alternation scenarios, we pre-train with TD3-norm or SAC-norm and then finetune from the resulting checkpoints using the corresponding streaming counterpart. We refer to \autoref{sec:supp_ft} for further details on the hyperparameters used during pre-training and finetuning, the checkpoints used for the experiments, and all details needed for reproducibility.

\begin{figure}[h]
    \begin{center}
        \includegraphics[width=1.0\textwidth]{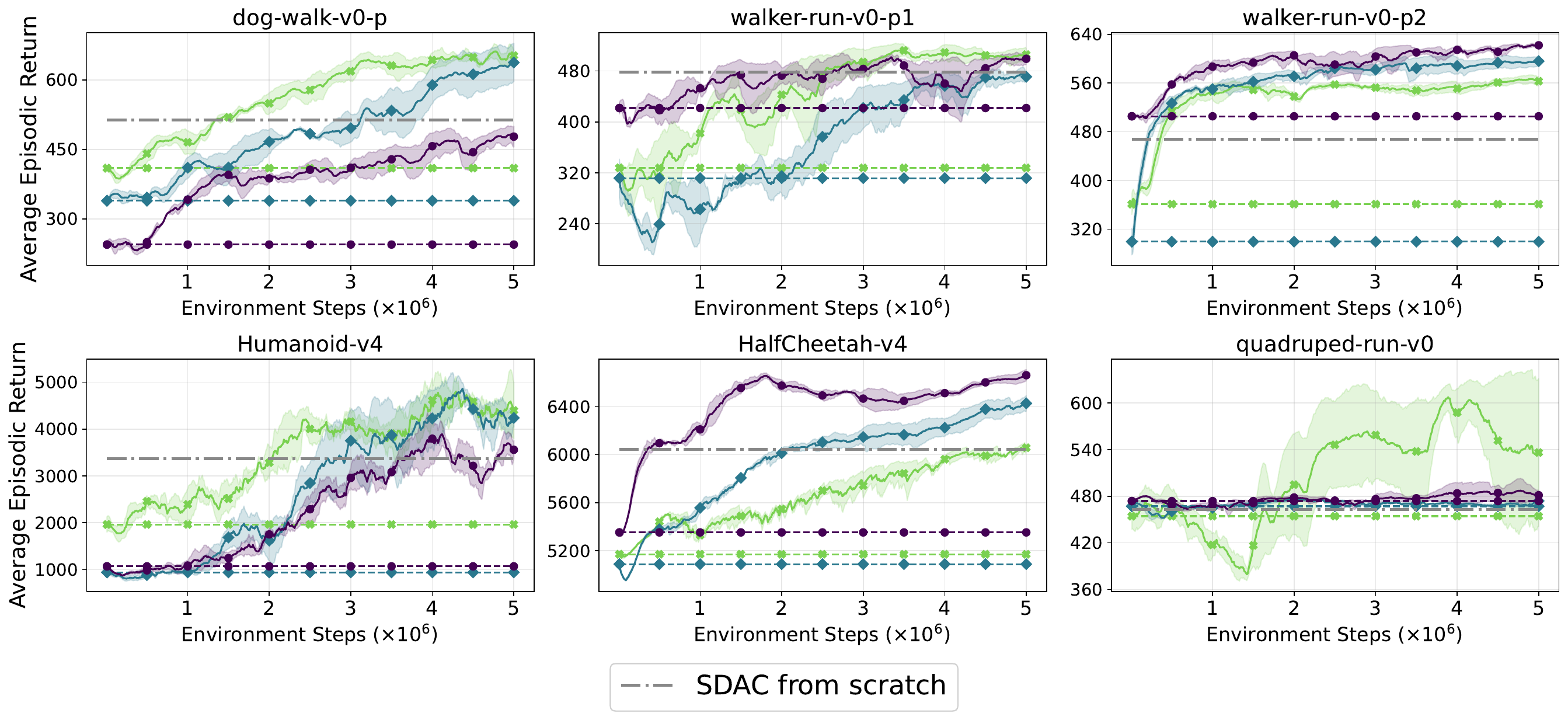}
    \end{center}
    \caption{Results for SDAC finetuned from
    TD3-norm checkpoints pre-trained with SGDC as the critic optimizer. \textbf{How to read.} Each of the three coloured curves is one pre-trained checkpoint, obtained from a different pre-training seed, finetuned with SDAC; the horizontal dashed line of the same colour marks the return of that checkpoint at the moment of the switch. A curve above its own dashed line therefore means finetuning has improved on the policy it inherited. The grey dash-dotted line is the return SDAC reaches when trained \emph{from scratch} in the same environment for 20M steps. Shaded bands are 95\% confidence intervals over three finetuning seeds per checkpoint.}
    \label{fig:ft_main}
\end{figure}

In the following experiments we evaluate finetuning performances
%\paragraph{Finetuning evaluation} The following experiments consider the finetuning
by use of SDAC for 5M steps given a checkpoint obtained by TD3-norm with SGDC for the critic. Experiments concerning SAC are instead reported in \autoref{sec:supp_sac_ft}.

\paragraph{Finetuning results}\autoref{fig:ft_main} shows the finetuning results when switching from TD3-norm, with SGDC
for the critic, to SDAC. Considering these together with the three additional pre-training
checkpoints per environment reported in \autoref{subsec:supp_ft_add_res}, that is six
checkpoints per environment, two main observations follow.

First, in five of the six environments the transferred agent ends finetuning above the
performance of the policy it started from, and the improvement is obtained within 5M
streaming steps. The exception is \texttt{quadruped-run-v0}, where most checkpoints remain
close to their starting performance for the whole 5M steps and one degrades at the beginning of finetuning.\\
Second, in the majority of cases the finetuned agent matches or exceeds within these 5M
steps the level that SDAC from scratch reaches only after 20M steps, a fourfold difference
in interaction budget. This is the comparison that matters for the deployment scenarios, where environment interaction happens on the physical system
and is the scarce resource.
% It is not uniform, however: on \texttt{dog-walk-v0-p},
% \texttt{walker-run-v0-p2} and \texttt{HalfCheetah-v4} individual checkpoints finish below
% the from-scratch reference.

% Third, and in contrast to the Adam pre-training of \autoref{subsec:naive}, the
% degradation that remains is transient rather than terminal. Several runs still dip below
% their initial performance during the first one to two million steps, most visibly on
% \texttt{walker-run-v0-p1}, and on \texttt{Humanoid-v4} and \texttt{HalfCheetah-v4} among
% the additional checkpoints, before recovering and improving. SGDC pre-training therefore
% does not eliminate the transition cost; what it removes is the regime in which the agent
% never recovers. For deployment this distinction matters, and it also bounds what we claim:
% the pipeline makes the transition survivable and productive, not free.
Results hold both in
the perturbed environments and in the unmodified ones, suggesting that pre-training with
TD3-norm and SGDC provides an initialization that transfers across both conditions.

\section{Conclusions}
\label{sec:conc}

Streaming deep reinforcement learning is usually evaluated as a self-contained regime, with
agents trained from scratch until convergence. The deployment scenarios that motivate the
paradigm, on-device adaptation under Sim2Real shift and falling back to a lighter learner when
compute must be shared with other subsystems, instead require carrying a batch-trained agent
\emph{into} streaming learning. We showed that this transition is neither automatic nor
cosmetic: handing a policy pre-trained with TD3 to a streaming learner can destroy the
performance that pre-training was meant to secure, even when the weights transfer exactly. We
traced the failure to the optimizers acting on either side of the switch, showed that ObGD
coincides with SGDC in the streaming regime, and used that identification to obtain a remedy
that leaves the streaming algorithm untouched.
% : pre-train the critic with SGDC, then warm up the
% critic before releasing the policy.
Under our pipeline, the transferred agent improves on the
policy it started from in five of six environments, in most cases reaching within 5M streaming
steps a level that streaming from scratch attains only after 20M. Making the transition
expressible in the first place required S2AC and SDAC, streaming counterparts of SAC and TD3
that are competitive with the state-of-the-art streaming baseline while sharing the objective
and topology of their batch counterparts; the ingredients that make streaming learning viable,
state normalization, reward scaling and LayerNorm, also improve SAC and TD3 substantially in
the batch regime. Batch and streaming algorithms therefore need not be treated as independent
paradigms, but may be designed around a shared foundation: shared architectures, shared
objectives and, as this work shows, shared optimizers.

\paragraph{Limitations} Two limitations mark where that foundation is still incomplete. First, the transition is not
uniformly successful. On \texttt{quadruped-run-v0} transfer neither degrades nor meaningfully
improves the pre-trained policy for most checkpoints, and one checkpoint degrades at the beginning of finetuning. Second, the maximum-entropy branch remains open: S2AC carries a
residual tuning burden that SDAC does not, as the entropy coefficient must be chosen per domain
(\autoref{subsec:streamingexp}), while on the batch side SAC-norm with SGDC does not yet train
reliably and finetuning is particularly sensitive to the entropy coefficient
(\autoref{sec:supp_sac_ft}). Maximum entropy RL nonetheless remains a promising direction for
finetuning, as entropy regularization naturally promotes robustness and behavioral diversity
under distribution shift \citep{eysenbach2022maximum}.

% \begin{ack}
% % Use unnumbered first level headings for the acknowledgments. All acknowledgments
% % go at the end of the paper before the list of references. Moreover, you are required to declare
% % funding (financial activities supporting the submitted work) and competing interests (related financial activities outside the submitted work).
% % More information about this disclosure can be found at: \url{https://neurips.cc/Conferences/2026/PaperInformation/FundingDisclosure}.

% % Do {\bf not} include this section in the anonymized submission, only in the final paper. You can use the
% % \texttt{ack} environment provided in the style file to automatically hide this section in the anonymized submission.
% \end{ack}

\bibliographystyle{plain}
\bibliography{main}

%%%%%%%%%%%%%%%%%%%%%%%%%%%%%%%%%%%%%%%%%%%%%%%%%%%%%%%%%%%%

\appendix
\newpage

\section{State Normalization and Reward Scaling}\label{sec:supp_norm}
We report here the pseudocodes for the NormalizeObservation (state normalization) and ScaleReward algorithms of Andrychowitz et al. \cite{andrychowicz2020matters} and Engstrom et al.  \cite{engstrom2020implementation} respectively, both of which exploit the online algorithm of Welford \cite{welford1962note} to track the required statistics. Without loss of generality, we simplify the notation by considering a scalar state $s$.

\begin{algorithm}
\setstretch{1.3}
\caption{SampleMeanVar \citep{welford1962note}}\label{alg:sample-mean-var}
\begin{algorithmic}[1]
\State \textbf{Given}: Input $x$, mean $\mu$, statistic $p$, and counter $n$.
\State $n \gets n + 1$
\State $\bar{\mu} \gets \mu + \frac{1}{n}(x-\mu)$
\State $p \gets p + (x-\mu)(x-\bar{\mu})$
\State $\sigma^2\gets \frac{p}{n-1}$ if $n\geq 2$, otherwise $\sigma^2\gets 1$
\State \textbf{return} $\bar{\mu}, p, \sigma^2, n$
\end{algorithmic}
\end{algorithm}

\begin{algorithm}
\setstretch{1.3}
\caption{ScaleReward \citep{engstrom2020implementation}}\label{alg:reward-norm}
\begin{algorithmic}[1]
\State \textbf{Initialize}: $u \leftarrow 0$
\State \textbf{Given:} $r, \gamma, p, T, n$
\State $u \leftarrow \gamma (1-T) u + r$
\State $\_, p,\sigma^2, n \leftarrow $ SampleMeanVar($u, 0, p, n$)
\State \textbf{return} $\frac{r}{\sqrt{\sigma^2 + \epsilon}}$, $p$
\end{algorithmic}
\end{algorithm}

\begin{algorithm}
\setstretch{1.3}
\caption{NormalizeObservation \citep{andrychowicz2020matters}}\label{alg:obs-norm}
\begin{algorithmic}[1]
\State \textbf{Given:} $s, \mu, p, n$
\State $\mu, \sigma^2, p, n \leftarrow $ SampleMeanVar($s, \mu, p, n$)
\State \textbf{return}  $\frac{s-\mu}{\sqrt{\sigma^2 + \epsilon}}$, $\mu$, $p$
\end{algorithmic}
\end{algorithm}

\newpage
\section{Pseudocode for S2AC and SDAC}\label{sec:supp_pseudocodes}

The pseudocodes for S2AC and SDAC are reported in Algorithm~\ref{alg:s2ac} 
and Algorithm~\ref{alg:sdac}, respectively, while the NormalizeObservation 
and ScaleReward subroutines are detailed in \autoref{sec:supp_norm}.

\begin{algorithm}
\caption{Streaming Soft Actor-Critic (S2AC)}
\label{alg:s2ac}
\begin{algorithmic}[1]

\State \textbf{Given} LayerNorm policy network $\pi_{\boldsymbol{\theta}}(\boldsymbol{a}|\boldsymbol{s})$ parameterizing a normal distribution with vectorized weights vector $\boldsymbol{\theta}$ and initialized with SparseInit
\State \textbf{Given} LayerNorm soft action-value network $Q^{\text{soft}}(\boldsymbol{s},\boldsymbol{a};\boldsymbol{\phi})$ with vectorized weights vector $\boldsymbol{\phi}$ and initialized with SparseInit
\State \textbf{Initialize} discount factor $\gamma$ and eligibility traces parameter $\lambda$ for the critic
\State \textbf{Initialize} policy learning rate $\eta_\pi$, action-value step size $\eta_{Q}$, starting entropy coefficient $\alpha_0$, and action-value scaling factor $\kappa_{Q}$
\State \textbf{Initialize} $p_r, p_s$ to zero and $\mu_s, t$ to one
\For{each episode}
    \State $\boldsymbol{z} \leftarrow \boldsymbol{0}$
    \State Initialize $\boldsymbol{s}$ (first state of the episode)
    \State $\boldsymbol{s}, \mu_s, p_s \leftarrow$ NormalizeObservation$(\boldsymbol{s}, \mu_s, p_s, t)$
    \For{each time step in the episode}
        \State $t \leftarrow t + 1$
        \State $\boldsymbol{a} \sim \pi_{\boldsymbol{\theta}}(\cdot|\boldsymbol{s})$
        \State Take action $\boldsymbol{a}$, observe $\boldsymbol{s}'$, $r$, $T$ \Comment{$T$ indicates whether $\boldsymbol{s}'$ is a terminal state}
        \State $\boldsymbol{s}', \mu_s, p_s \leftarrow$ NormalizeObservation$(\boldsymbol{s}', \mu_s, p_s, t)$
        \State $r, p_r, \sigma_r \leftarrow$ ScaleReward$(r, \gamma, p_r, T, t)$
        \State $\alpha \leftarrow \alpha_0/\sigma_r$
        \State $\boldsymbol{a}' \sim \pi_{\boldsymbol{\theta}}(\cdot|\boldsymbol{s}')$
        \State $\delta \leftarrow r + \gamma (1-T) (Q^{\text{soft}}(\boldsymbol{s}',\boldsymbol{a}'; \boldsymbol{\phi}) - \alpha \log\pi_{\boldsymbol{\theta}}(\boldsymbol{a}'|\boldsymbol{s}')) - Q^{\text{soft}}(\boldsymbol{s},\boldsymbol{a}; \boldsymbol{\phi})$
        \State $\boldsymbol{z} \leftarrow \gamma\lambda\boldsymbol{z} + \nabla_{\boldsymbol{\phi}}Q^{\text{soft}}(\boldsymbol{s},\boldsymbol{a}; \boldsymbol{\phi})$
        \State $J_\pi(\boldsymbol{\theta}) \leftarrow \alpha\log\pi_{\boldsymbol{\theta}}(\boldsymbol{a}|\boldsymbol{s}) - Q^{\text{soft}}(\boldsymbol{s},\boldsymbol{a};\boldsymbol{\phi})$ 
        \State $\boldsymbol{\phi} \leftarrow$ ObGD$(\boldsymbol{z}, \boldsymbol{\phi}, \delta, \eta_{Q}, \kappa_{Q})$
        \State $\boldsymbol{\theta} \leftarrow \boldsymbol{\theta} - \eta_\pi \nabla_{\boldsymbol{\theta}}J_\pi(\boldsymbol{\theta})$ \Comment{Update performed with Adam}
        \State $\boldsymbol{s} \leftarrow \boldsymbol{s}'$
    \EndFor
\EndFor

\end{algorithmic}
\end{algorithm}

\begin{algorithm}
\caption{Streaming Deterministic Actor-Critic (SDAC)}
\label{alg:sdac}
\begin{algorithmic}[1]

\State \textbf{Given} LayerNorm actor network $\pi_{\boldsymbol{\theta}}(\boldsymbol{s})$ parameterizing a deterministic policy with vectorized weights vector $\boldsymbol{\theta}$ and initialized with SparseInit
\State \textbf{Given} LayerNorm action-value network $Q(\boldsymbol{s},\boldsymbol{a};\boldsymbol{\phi})$ with vectorized weights vector $\boldsymbol{\phi}$ and initialized with SparseInit
\State \textbf{Initialize} discount factor $\gamma$ and eligibility traces parameter $\lambda$ for the critic
\State \textbf{Initialize} policy learning rate $\eta_\pi$, action-value step size $\eta_{Q}$, Gaussian std $\sigma$, and action-value scaling factor $\kappa_{Q}$
\State \textbf{Initialize} $p_r, p_s$ to zero and $\mu_s, t$ to one
\For{each episode}
    \State $\boldsymbol{z} \leftarrow \boldsymbol{0}$
    \State Initialize $\boldsymbol{s}$ (first state of the episode)
    \State $\boldsymbol{s}, \mu_s, p_s \leftarrow$ NormalizeObservation$(\boldsymbol{s}, \mu_s, p_s, t)$
    \For{each time step in the episode}
        \State $t \leftarrow t + 1$
        \State $\boldsymbol{\epsilon}_1\sim \mathcal{N}(\boldsymbol{0},\boldsymbol{I}\sigma^2)$
        \State $\boldsymbol{a} \leftarrow \pi_{\boldsymbol{\theta}}(\boldsymbol{s}) + \boldsymbol{\epsilon}_1$
        \State Take action $\boldsymbol{a}$, observe $\boldsymbol{s}'$, $r$, $T$ \Comment{$T$ indicates whether $\boldsymbol{s}'$ is a terminal state}
        \State $\boldsymbol{s}', \mu_s, p_s \leftarrow$ NormalizeObservation$(\boldsymbol{s}', \mu_s, p_s, t)$
        \State $r, p_r \leftarrow$ ScaleReward$(r, \gamma, p_r, T, t)$
        \State $\boldsymbol{\epsilon}_2\sim \mathcal{N}(\boldsymbol{0},\boldsymbol{I}\sigma^2)$
        \State $\boldsymbol{a}' \leftarrow \pi_{\boldsymbol{\theta}}(\boldsymbol{s}')+\boldsymbol{\epsilon}_2$
        \State $\delta \leftarrow r + \gamma (1-T) Q(\boldsymbol{s}',\boldsymbol{a}'; \boldsymbol{\phi}) - Q(\boldsymbol{s},\boldsymbol{a}; \boldsymbol{\phi})$
        \State $\boldsymbol{z} \leftarrow \gamma\lambda\boldsymbol{z} + \nabla_{\boldsymbol{\phi}}Q(\boldsymbol{s},\boldsymbol{a}; \boldsymbol{\phi})$
        \State $J_\pi(\boldsymbol{\theta}) \leftarrow - Q(\boldsymbol{s},\pi_{\boldsymbol{\theta}}(\boldsymbol{s});\boldsymbol{\phi})$ 
        \State $\boldsymbol{\phi} \leftarrow$ ObGD$(\boldsymbol{z}, \boldsymbol{\phi}, \delta, \eta_{Q}, \kappa_{Q})$
        \State $\boldsymbol{\theta} \leftarrow \boldsymbol{\theta} - \eta_\pi \nabla_{\boldsymbol{\theta}}J_\pi(\boldsymbol{\theta})$ \Comment{Update performed with Adam}
        \State $\boldsymbol{s} \leftarrow \boldsymbol{s}'$
    \EndFor
\EndFor

\end{algorithmic}
\end{algorithm}

\clearpage
\section{Hyperparameters and network architectures for S2AC, SDAC and Stream AC($\lambda$)}\label{sec:supp_stream_param}

\subsection{Hyperparameters}
\label{subsec:supp_stream_hyper}

Regardless of the streaming algorithm and environment, we set $\gamma=0.99$. For the hyperparameters of Stream AC$(\lambda)$ we follow Elsayed et al.\ \cite{elsayed2024streaming}, and \autoref{table:StreamAC_hyper} reports all values. \autoref{table:SDAC_hyper} reports the hyperparameters concerning SDAC, while \autoref{table:S2AC_hyper} the ones for S2AC. Concerning the base entropy coefficient $\alpha_0$ of S2AC, \autoref{table:s2ac_alpha} reports the specific value per environment.

\begin{table}[h]
\caption{Stream AC($\lambda$) Hyperparameters.}\label{table:StreamAC_hyper}
\centering
\begin{tabular}{cll}
\toprule
& Hyperparameter & Value \\
\midrule
\multirow{2}{*}{}
& entropy coefficient $\tau$      & $0.01$ \\
\midrule
\multirow{4}{*}{Actor Optimizer} 
& Optimizer        & ObGD~\citep{elsayed2024streaming} \\
& Learning rate    & $1.0$ \\
& $\lambda$    & $0.8$ \\
& $\kappa$    & $3.0$ \\
\midrule
\multirow{4}{*}{Critic Optimizer} 
& Optimizer        & ObGD~\citep{elsayed2024streaming} \\
& Learning rate    & $1.0$ \\
& $\lambda$    & $0.8$ \\
& $\kappa$    & $2.0$ \\
\bottomrule
\end{tabular}
\end{table}

\begin{table}[h]
\caption{SDAC Hyperparameters.}\label{table:SDAC_hyper}
\centering
\begin{tabular}{cll}
\toprule
& Hyperparameter & Value \\
\midrule
\multirow{2}{*}{}
& Target Noise      & $\mathcal{N}(0,0.2^2)$ \\
& Exploration noise    & $\mathcal{N}(0,0.2^2)$ \\
\midrule
\multirow{3}{*}{Actor Optimizer} 
& Optimizer        & Adam~\citep{kingma2014adam} \\
& Learning rate    & $3\text{e}-4$ \\
& Betas            & $\beta_1=0.9, \beta_2=0.999$\\
\midrule
\multirow{4}{*}{Critic Optimizer} 
& Optimizer        & ObGD~\citep{elsayed2024streaming} \\
& Learning rate    & $1.0$ \\
& $\lambda$    & $0.8$ \\
& $\kappa$    & $2.0$ \\
\bottomrule
\end{tabular}
\end{table}

\begin{table}[h]
\caption{S2AC Hyperparameters.}\label{table:S2AC_hyper}
\centering
\begin{tabular}{cll}
\toprule
& Hyperparameter & Value \\
\midrule
\multirow{2}{*}{}
& log std max      & $2$ \\
& log std min    & $-20$ \\
\midrule
\multirow{3}{*}{Actor Optimizer} 
& Optimizer        & Adam~\citep{kingma2014adam} \\
& Learning rate    & $3\text{e}-4$ \\
& Betas            & $\beta_1=0.9, \beta_2=0.999$\\
\midrule
\multirow{4}{*}{Critic Optimizer} 
& Optimizer        & ObGD~\citep{elsayed2024streaming} \\
& Learning rate    & $1.0$ \\
& $\lambda$    & $0.8$ \\
& $\kappa$    & $2.0$ \\
\bottomrule
\end{tabular}
\end{table}

\begin{table}[h]
\caption{Base entropy coefficient $\alpha_0$ of S2AC, per environment.}\label{table:s2ac_alpha}
\centering
\begin{tabular}{clc}
\toprule
&Environment & $\alpha_0$ \\
\midrule
\multirow{6}{*}{MuJoCo Gym \citep{todorov2012mujoco}}&Humanoid-v4 & \multirow{6}{*}{$0.2$}\\
&HumanoidStandup-v4 &\\
&HalfCheetah-v4 &\\
&Hopper-v4 &\\
&Walker2d-v4&\\
&Ant-v4&\\
\midrule
\multirow{8}{*}{DM Control Suite \citep{tunyasuvunakool2020dm_control}}&dog-walk-v0 &\multirow{4}{*}{$0.002$}\\
&cheetah-run-v0&\\
&walker-run-v0&\\
&quadruped-run-v0&\\
\cmidrule{2-3}
&cartpole-swingup-v0&0.01\\
\cmidrule{2-3}
&finger-spin-v0&0.012\\
\bottomrule
\end{tabular}
\end{table}

% Keep the hyperparameter tables from floating past this point: without the
% barrier they migrate into the following subsection and interleave with the
% architecture boxes, which are not floats.
%\FloatBarrier
\clearpage
\subsection{Network architectures}
\label{subsec:supp_stream_arch}

Following Elsayed et al.\ \cite{elsayed2024streaming}, the same architecture is used for both the critic and the actor. Pseudocode \ref{pseudo:StreamAC}, Pseudocode \ref{pseudo:SDAC} and Pseudocode \ref{pseudo:S2AC} report it for each streaming algorithm.

Following Haarnoja et al.\ \cite{haarnoja2018soft}, we apply squashing for S2AC: denoting with $\mu_{\boldsymbol{\theta}}(\cdot|\boldsymbol{s})$ the Gaussian distribution parametrized by the actor network's output and by $\boldsymbol{u}\in\mathbb{R}^D$ a sample from it, we enforce action bounds via $\boldsymbol{a}=\tanh(\boldsymbol{u})$. The corresponding log-likelihood is:

\begin{equation}
    \log \pi_{\boldsymbol{\theta}}(\boldsymbol{a}|\boldsymbol{s})=\sum_{i=1}^D\log \mu_{\boldsymbol{\theta}}(u_i|\boldsymbol{s})-2\cdot (\log 2-u_i-\text{SoftPlus}(-2\cdot u_i ))
\end{equation}

where $u_i$ denotes the $i$-th component of $\boldsymbol{u}$.

\begin{tcolorbox}[title={\begin{pseudo}\label{pseudo:StreamAC}Stream AC$(\lambda)$ Network Details\end{pseudo}}]
\textbf{Value $V$ Network:}

\begin{verbatim}
fc_layer = Linear(state_dim, 128)
hidden_layer = Linear(128, 128)
output_linear = Linear(128, 1)
\end{verbatim} 

\textbf{Value $V$ Forward Pass:}
\begin{verbatim}
input = state
x = LeakyReLU(layer_norm(fc_layer(input)))
x = LeakyReLU(layer_norm(hidden_layer(x)))

value = output_linear(x)
\end{verbatim} 

\vspace{-8pt}
\hrulefill

\textbf{Policy $\pi$ Network:}
\begin{verbatim}
fc_layer = Linear(state_dim, 128)
hidden_layer = Linear(128, 128)

linear_mu = Linear(128, action_dim)
linear_std = Linear(128, action_dim)
\end{verbatim} 

\textbf{Policy $\pi$ Forward Pass:}
\begin{verbatim}
input = state
x = LeakyReLU(layer_norm(fc_layer(input)))
x = LeakyReLU(layer_norm(hidden_layer(x)))

mu = linear_mu(x)
pre_std = linear_std(x)
std = SoftPlus(pre_std)

action = Normal(mu, std).sample()
\end{verbatim} 
\end{tcolorbox}

\begin{tcolorbox}[title={\begin{pseudo}\label{pseudo:SDAC}SDAC Network Details\end{pseudo}}]
\textbf{Value $Q$ Network:}

\begin{verbatim}
fc_layer = Linear(state_dim + action_dim, 128)
hidden_layer = Linear(128, 128)
output_linear = Linear(128, 1)
\end{verbatim} 

\textbf{Value $Q$ Forward Pass:}
\begin{verbatim}
input = concatenate([state, action])
x = LeakyReLU(layer_norm(fc_layer(input)))
x = LeakyReLU(layer_norm(hidden_layer(x)))

value = output_linear(x)
\end{verbatim} 

\vspace{-8pt}
\hrulefill

\textbf{Policy $\pi$ Network:}
\begin{verbatim}
fc_layer = Linear(state_dim, 128)
hidden_layer = Linear(128, 128)

linear_mu = Linear(128, action_dim)
\end{verbatim} 

\textbf{Policy $\pi$ Forward Pass:}
\begin{verbatim}
input = state
x = LeakyReLU(layer_norm(fc_layer(input)))
x = LeakyReLU(layer_norm(hidden_layer(x)))

action = tanh(linear_mu(x))
\end{verbatim} 
\end{tcolorbox}

\begin{tcolorbox}[title={\begin{pseudo}\label{pseudo:S2AC}S2AC Network Details\end{pseudo}}]
\textbf{Value $Q^{\text{soft}}$ Network:}

\begin{verbatim}
fc_layer = Linear(state_dim + action_dim, 128)
hidden_layer = Linear(128, 128)
output_linear = Linear(128, 1)
\end{verbatim} 

\textbf{Value $Q^{\text{soft}}$ Forward Pass:}
\begin{verbatim}
input = concatenate([state, action])
x = LeakyReLU(layer_norm(fc_layer(input)))
x = LeakyReLU(layer_norm(hidden_layer(x)))

value = output_linear(x)
\end{verbatim} 

\vspace{-8pt}
\hrulefill

\textbf{Policy $\pi$ Network:}
\begin{verbatim}
fc_layer = Linear(state_dim, 128)
hidden_layer = Linear(128, 128)
linear_mu = Linear(128, action_dim)
linear_std = Linear(128, action_dim)

LOG_STD_MAX = 2
LOG_STD_MIN = -20
\end{verbatim} 

\textbf{Policy $\pi$ Forward Pass:}
\begin{verbatim}
input = state
x = LeakyReLU(layer_norm(fc_layer(input)))
x = LeakyReLU(layer_norm(hidden_layer(x)))

mu = linear_mu(x)
log_std = linear_std(x)

log_std = tanh(log_std)
log_std = LOG_STD_MIN + \
0.5 * (LOG_STD_MAX - LOG_STD_MIN) * (log_std + 1)

x = Normal(mu, exp(log_std)).sample()
action = tanh(x)
\end{verbatim} 
\end{tcolorbox}

\clearpage
\section{Additional results for streaming RL}\label{sec:supp_stream_res}

\begin{figure}[h]
    \begin{center}
        \includegraphics[width=1.0\textwidth]{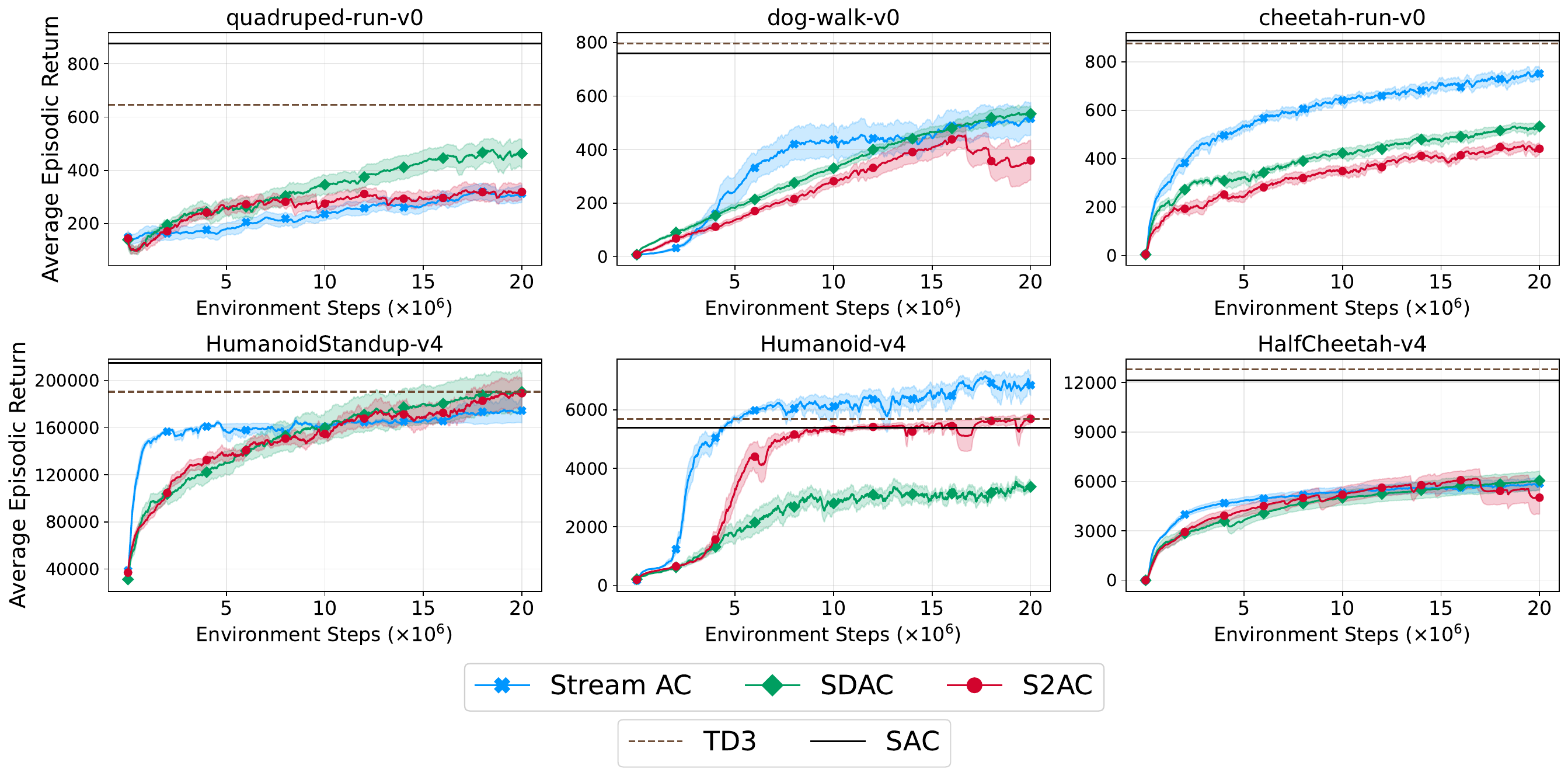}
    \end{center}
    \caption{Results for streaming DRL algorithms SDAC, S2AC, and Stream AC$(\lambda)$ on MuJoCo Gym and DM Control Suite tasks. TD3 and SAC in the legend denote results with state normalization and reward scaling, trained over 3M steps.}
    \label{fig:stream_supp_1}
\end{figure}

\begin{figure}[h]
    \begin{center}
        \includegraphics[width=1.0\textwidth]{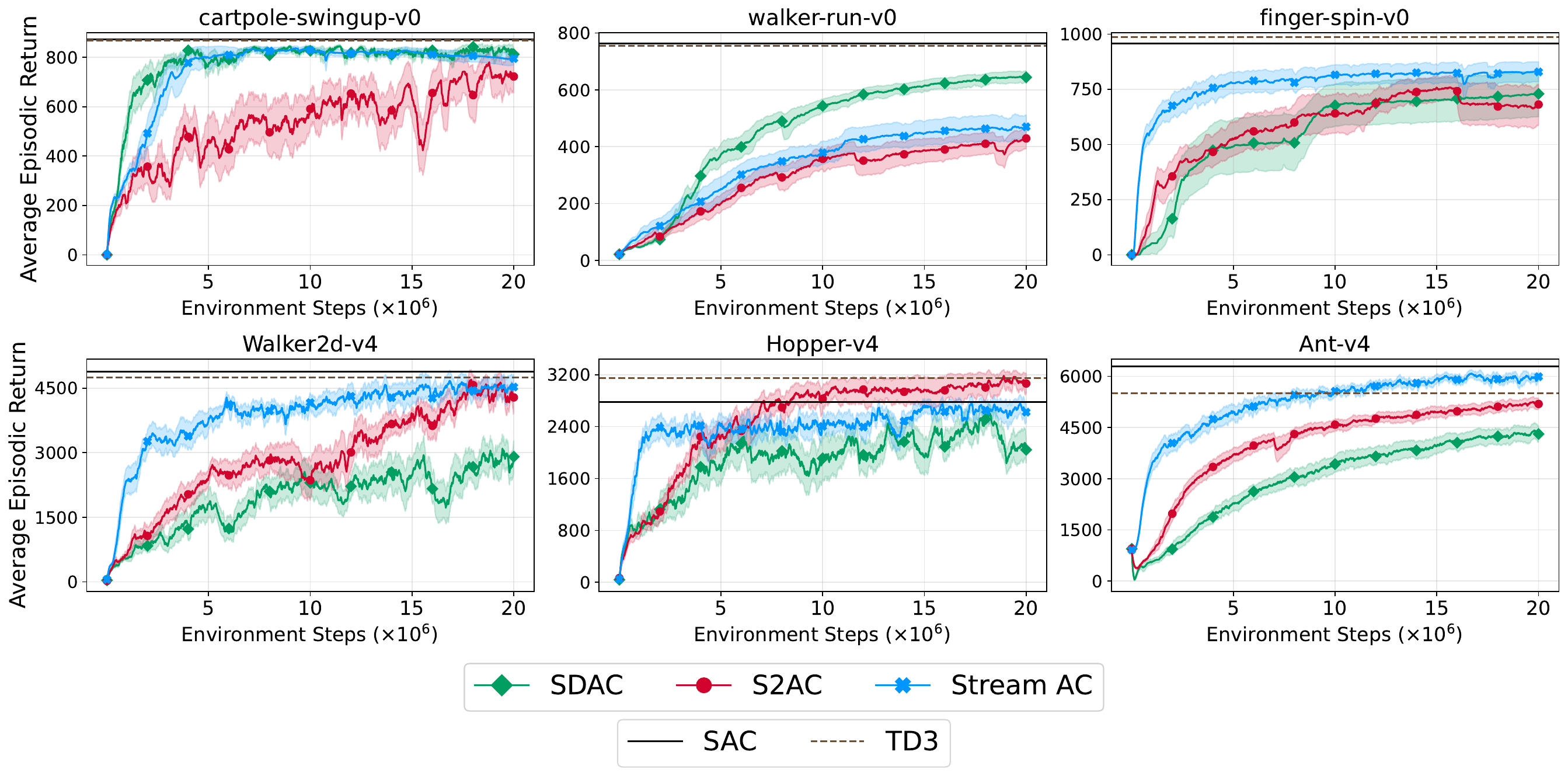}
    \end{center}
    \caption{Additional results for streaming DRL algorithms SDAC, S2AC, and Stream AC$(\lambda)$ on MuJoCo Gym and DM Control Suite tasks. TD3 and SAC in the legend denote results with state normalization and reward scaling, trained over 3M steps.}
    \label{fig:stream_supp_2_3}
\end{figure}

We present additional results for the streaming algorithms. In \autoref{fig:stream_supp_1} we present the same results as in \autoref{subsec:streamingexp} including a comparison with TD3 and SAC, while \autoref{fig:stream_supp_2_3} presents additional results on different environments from DM Control Suite and MuJoCo Gym.

\clearpage
\section{Comparison with AVG}
\label{sec:supp_avg}

AVG \citep{vasan2024deep} is a streaming actor-critic for continuous action control based on
the maximum-entropy RL framework and, like S2AC, can be regarded as a streaming counterpart of
SAC \citep{haarnoja2018soft}, which makes it the most directly comparable baseline for our
evaluation. Two constraints limit the scope of the comparison. First, AVG requires extensive
per-environment hyperparameter tuning, which makes running it over the full benchmark suite
intractable. Second, Vasan et al.\ \cite{vasan2024deep} evaluate AVG only on MuJoCo Gym
\citep{todorov2012mujoco} and on a restricted subset of DM Control Suite
\citep{tunyasuvunakool2020dm_control} environments, namely the Dog and Finger Spin domains,
excluding others such as Quadruped and Walker. We therefore restrict the comparison to the
environments considered in the original paper, following the experimental setup of
Vasan et al.\ \cite{vasan2024deep} with the exception of the hidden layer size, which we reduce to 128 to
match the other streaming approaches.

\begin{figure}[h]
    \begin{center}
        \includegraphics[width=1.0\textwidth]{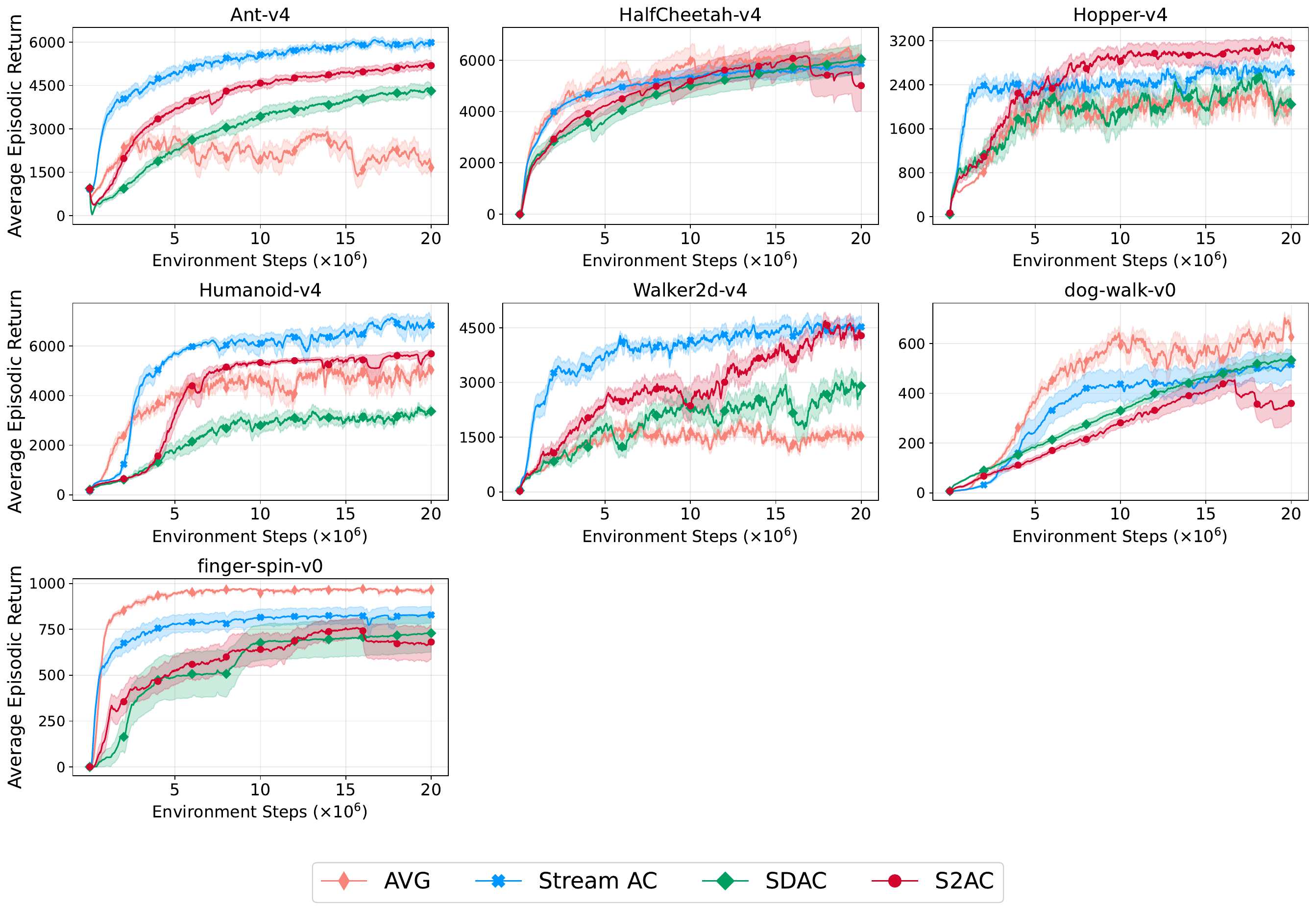}
    \end{center}
    \caption{Results for streaming DRL algorithms SDAC, S2AC, Stream AC$(\lambda)$, and AVG on MuJoCo Gym and DM Control Suite tasks.}
    \label{fig:avg}
\end{figure}

The results in \autoref{fig:avg} reveal a highly inconsistent performance profile for AVG:
while it achieves strong results on some environments such as \texttt{dog-walk-v0}, it
performs poorly on others such as \texttt{Ant-v4}. This variability holds despite
per-environment tuning of learning rates, discount factor $\gamma$, and Adam
\citep{kingma2014adam} betas. S2AC tunes none of these and attains comparable overall
performance across the evaluated environments.

\clearpage
\newpage
\section{Ablation studies for SDAC and S2AC}
\label{sec:supp_abl_stream}
\begin{figure}[h]
    \begin{center}
        \includegraphics[width=\textwidth]{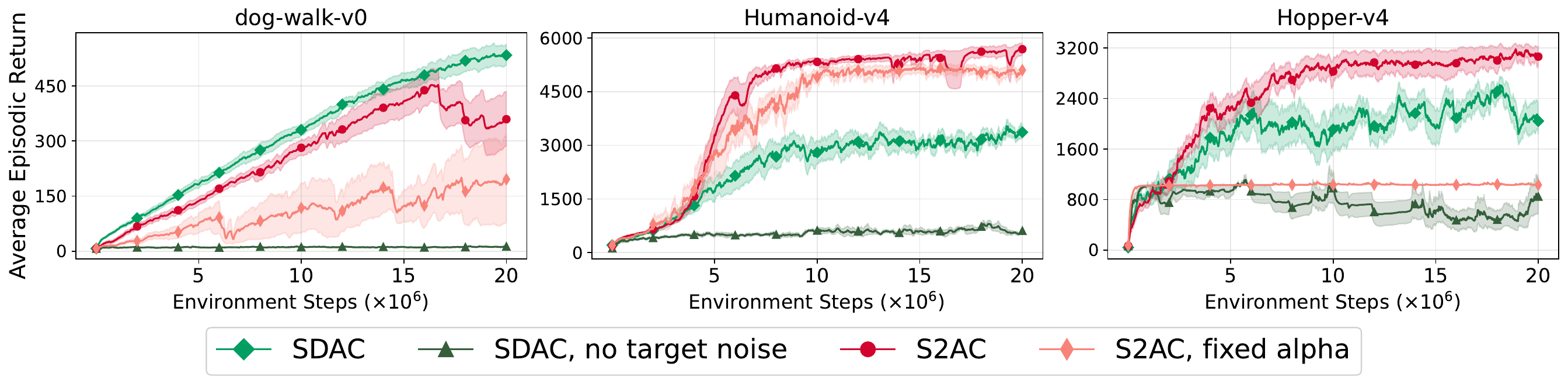}
    \end{center}
    \caption{Ablation for SDAC and S2AC: the adaptive entropy schedule $\alpha \leftarrow \alpha_0/\sigma_r$ for S2AC, and target noise for SDAC.}
    \label{fig:stream_ablation}
\end{figure}

\autoref{fig:stream_ablation} presents an ablation study examining the contribution of the adaptive schedule $\alpha/\sigma_r$ and the target noise proposed in \autoref{sec:method}. 
% In particular, for S2AC, we compare a fixed entropy coefficient $\alpha$ against the adaptive schedule $\alpha/\sigma_r$, while for SDAC, we investigate the role of target noise in \autoref{eq:tderror}. 
While S2AC still achieves decent performance without the adaptive schedule, the target noise modification is far more critical for SDAC, which fails to learn entirely without it.

\begin{figure}[ht]
    \begin{center}
        \includegraphics[width=1.0\textwidth]{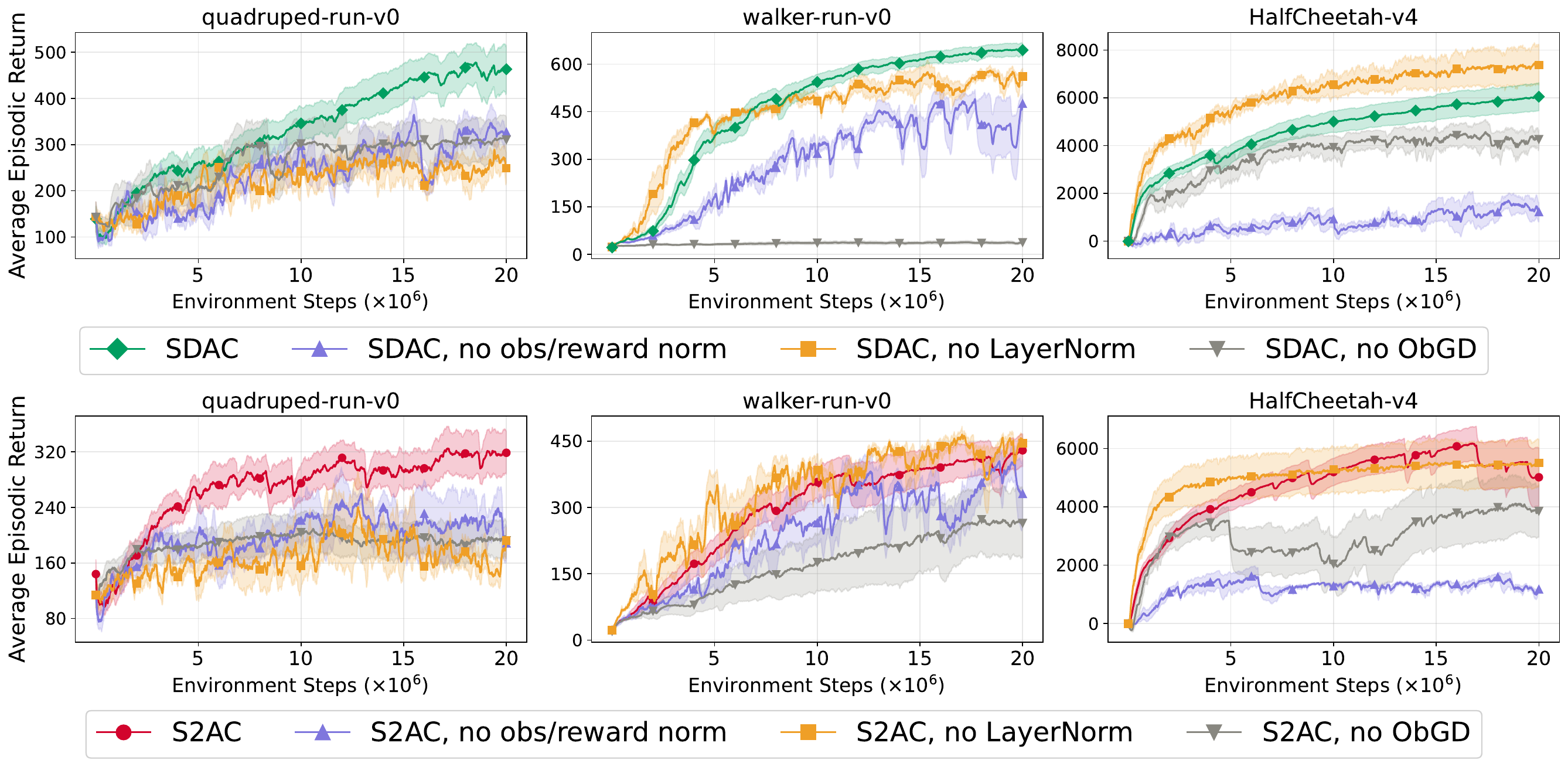}
    \end{center}
    \caption{Additional ablation studies for SDAC and S2AC.}
    \label{fig:ablation_sdac}
\end{figure}

Furthermore, we report additional ablation studies on both SDAC and S2AC, considering the removal of observation and reward normalization, the replacement of ObGD with Adam, and the removal of LayerNorm from the network architectures. As shown in Figure~\ref{fig:ablation_sdac}, the removal of ObGD is consistently the most damaging modification for SDAC, causing near-complete failure in the \texttt{walker-run-v0} environment, while its impact on S2AC is more variable across tasks. Removing LayerNorm yields competitive performance in some environments for both algorithms, suggesting that its benefit may be task-dependent. Finally, observation and reward normalization provide a consistent benefit both for S2AC and SDAC, confirming the findings of Elsayed et al. \cite{elsayed2024streaming}.

\clearpage
\section{Hyperparameters and network architectures for SAC and TD3}\label{sec:supp_batch_params}

For the SAC and TD3 implementations, we build upon the CleanRL repository 
\citep{huang2022cleanrl}. All the models were trained on a server equipped with 2 AMD EPYC 9224 CPUs (48 cores, 96 threads), 1.5 TB of RAM, and 6 NVIDIA L40S GPUs (46 GB VRAM each). The network architectures are reported in Pseudocode~\ref{pseudo:SAC} and Pseudocode~\ref{pseudo:TD3}, where the hidden layer size is reduced from 256 to 128 to match the capacity of the streaming approaches. The hyperparameters for SAC and TD3 follow those of the original works of Haarnoja et al. \cite{haarnoja2018soft} and Fujimoto et al. \cite{fujimoto2018addressing}, and are reported in \autoref{table:SAC_hyper} and \autoref{table:TD3_hyper}, respectively.

\noindent For SAC-norm, the same network architectures of S2AC (Pseudocode \ref{pseudo:S2AC}) are employed and the same hyperparameters of SAC are kept (\autoref{table:SAC_hyper}). Similarly, TD3-norm uses the same architectures of SDAC (Pseudocode \ref{pseudo:SDAC}) and the same hyperparameters of TD3 (\autoref{table:TD3_hyper}).

\begin{table}[h]
\caption{SAC and SAC-norm Hyperparameters.}\label{table:SAC_hyper}
\centering
\begin{tabular}{cll}
\toprule
& Hyperparameter & Value \\
\midrule
\multirow{4}{*}{No streaming params}
& Buffer size & $10^6$\\
&Batch size & $256$\\
&Exploration time steps & $5\times 10^3$\\
&Target smoothing coeff. & $\tau=0.005$\\
\midrule

\multirow{5}{*}{}
& Frequency policy update & $2$\\
& $\alpha$ & Autotune\\
& Target entropy & $-$action\_dim\\
& log std max      & $2$ \\
& log std min    & $-20$ \\
\midrule
\multirow{3}{*}{Optimizer} 
& Optimizer        & Adam~\citep{kingma2014adam} \\
& Learning rate    & $3\text{e}-4$ \\
& Betas            & $\beta_1=0.9, \beta_2=0.999$\\
\bottomrule
\end{tabular}
\end{table}

\begin{table}[h]
\caption{TD3 and TD3-norm Hyperparameters.}\label{table:TD3_hyper}
\centering
\begin{tabular}{cll}
\toprule
& Hyperparameter & Value \\
\midrule
\multirow{4}{*}{No streaming params}
& Buffer size & $10^6$\\
&Batch size & $256$\\
&Exploration time steps & $25\times 10^3$\\
&Target smoothing coeff. & $\tau=0.005$\\
\midrule

\multirow{5}{*}{}
& Frequency policy update & $2$\\
& Target Noise & $\mathcal{N}(0,0.2^2)$\\
&Exploration Noise & $\mathcal{N}(0,0.1^2)$\\
\midrule
\multirow{3}{*}{Optimizer} 
& Optimizer        & Adam~\citep{kingma2014adam} \\
& Learning rate    & $3\text{e}-4$ \\
& Betas            & $\beta_1=0.9, \beta_2=0.999$\\
\bottomrule
\end{tabular}
\end{table}

% See the note in stream_supp.tex: the barrier stops these floats from drifting
% into the architecture boxes below, which are ordinary text rather than floats.
\clearpage

\begin{tcolorbox}[title={\begin{pseudo}\label{pseudo:SAC}SAC Network Details\end{pseudo}}]
\textbf{Value $Q^{\text{soft}}$ Network:}
 
\begin{verbatim}
fc_layer = Linear(state_dim + action_dim, 128)
hidden_layer = Linear(128, 128)
output_linear = Linear(128, 1)
\end{verbatim} 

\textbf{Value $Q^{\text{soft}}$ Forward Pass:}
\begin{verbatim}
input = concatenate([state, action])
x = ReLU(fc_layer(input))
x = ReLU(hidden_layer(x))

value = output_linear(x)
\end{verbatim} 

\vspace{-8pt}
\hrulefill

\textbf{Policy $\pi$ Network:}
\begin{verbatim}
fc_layer = Linear(state_dim, 128)
hidden_layer = Linear(128, 128)
linear_mu = Linear(128, action_dim)
linear_std = Linear(128, action_dim)

LOG_STD_MAX = 2
LOG_STD_MIN = -20
\end{verbatim} 

\textbf{Policy $\pi$ Forward Pass:}
\begin{verbatim}
input = state
x = ReLU(fc_layer(input))
x = ReLU(hidden_layer(x))

mu = linear_mu(x)
log_std = linear_std(x)

log_std = tanh(log_std)
log_std = LOG_STD_MIN + \
0.5 * (LOG_STD_MAX - LOG_STD_MIN) * (log_std + 1)

x = Normal(mu, exp(log_std)).sample()
action = tanh(x)
\end{verbatim} 
\end{tcolorbox}

\begin{tcolorbox}[title={\begin{pseudo}\label{pseudo:TD3}TD3 Network Details\end{pseudo}}]
\textbf{Value $Q$ Network:}

\begin{verbatim}
fc_layer = Linear(state_dim + action_dim, 128)
hidden_layer = Linear(128, 128)
output_linear = Linear(128, 1)
\end{verbatim} 

\textbf{Value $Q$ Forward Pass:}
\begin{verbatim}
input = concatenate([state, action])
x = ReLU(fc_layer(input))
x = ReLU((hidden_layer(x))

value = output_linear(x)
\end{verbatim} 

\vspace{-8pt}
\hrulefill

\textbf{Policy $\pi$ Network:}
\begin{verbatim}
fc_layer = Linear(state_dim, 128)
hidden_layer = Linear(128, 128)

linear_mu = Linear(128, action_dim)
\end{verbatim} 

\textbf{Policy $\pi$ Forward Pass:}
\begin{verbatim}
input = state
x = ReLU(fc_layer(input))
x = ReLU(hidden_layer(x))

action = tanh(linear_mu(x))
\end{verbatim} 
\end{tcolorbox}

\clearpage
\section{Batch Deep RL results}
\label{sec:batch_supp}

\subsection{Data normalization for TD3 and SAC}
\label{subsec:supp_batch_norm}
\begin{figure}[h]
    \begin{center}
        \includegraphics[width=1.0\textwidth]{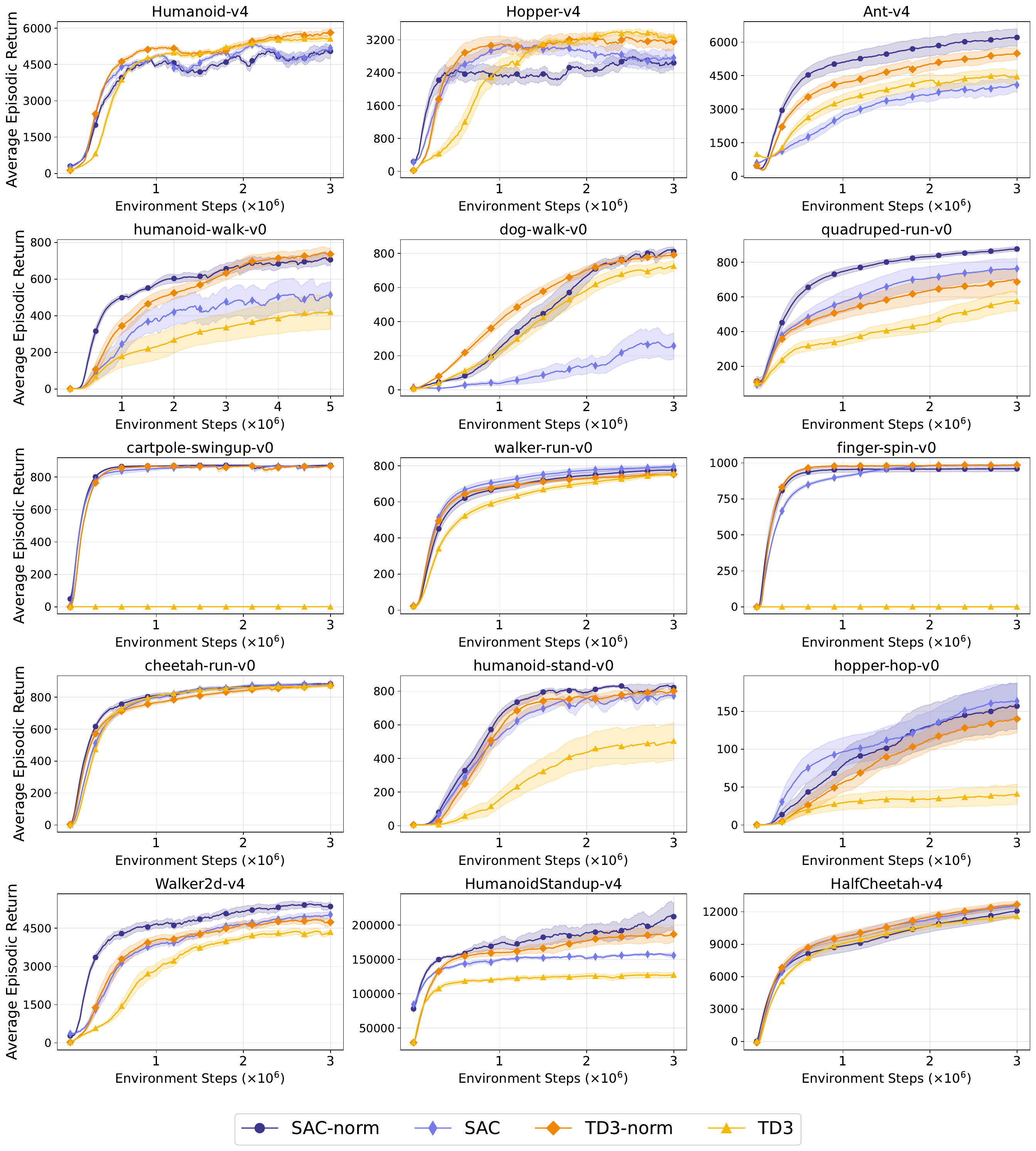}
    \end{center}
    \caption{Result for the batch methods on MuJoCo Gym and DM Control Suite tasks. TD3-norm and SAC-norm denote the versions of TD3 and SAC with data normalization and the same network architectures used for the streaming approaches.}
    \label{fig:batch_supp_1}
\end{figure}

Here, we investigate the effect of incorporating state normalization and reward scaling into batch methods, as described in \autoref{sec:b2s}. We refer to \autoref{sec:supp_batch_params} for details on the hyperparameters and network architectures used in the experiments.
%State normalization has been explored previously by \citet{lee2024simba} for the SimBa architecture, where it was combined with deeper residual networks \citep{he2016deep} and layer normalization to yield substantial performance gains for SAC and DDPG.
%In contrast, we consider a simple two-layer network with fewer hidden units (128 rather than 512 for the critic) and no residual connections, and extend the analysis to TD3.
As shown in \autoref{fig:batch_supp_1}, the architectures proposed by Elsayed et al.\ \citep{elsayed2024streaming} combined with data normalization not only preserve performance but yield notable improvements across many environments, particularly for TD3.
Beyond its role in the pipeline, this result is of independent interest: gains of the magnitude
reported by Lee et. al. \citep{lee2024simba} are attainable with a two-layer, 128-unit multilayer
perceptron, which matters when the same network must later run on the deployment device.

%Here, TD3 without normalization is not even able to learn, while normalization solves this.

\subsection{Pre-training the critic with SGDC}
\label{subsec:supp_batch_mix}

\begin{figure}[h]
    \begin{center}
        \includegraphics[width=1.0\textwidth]{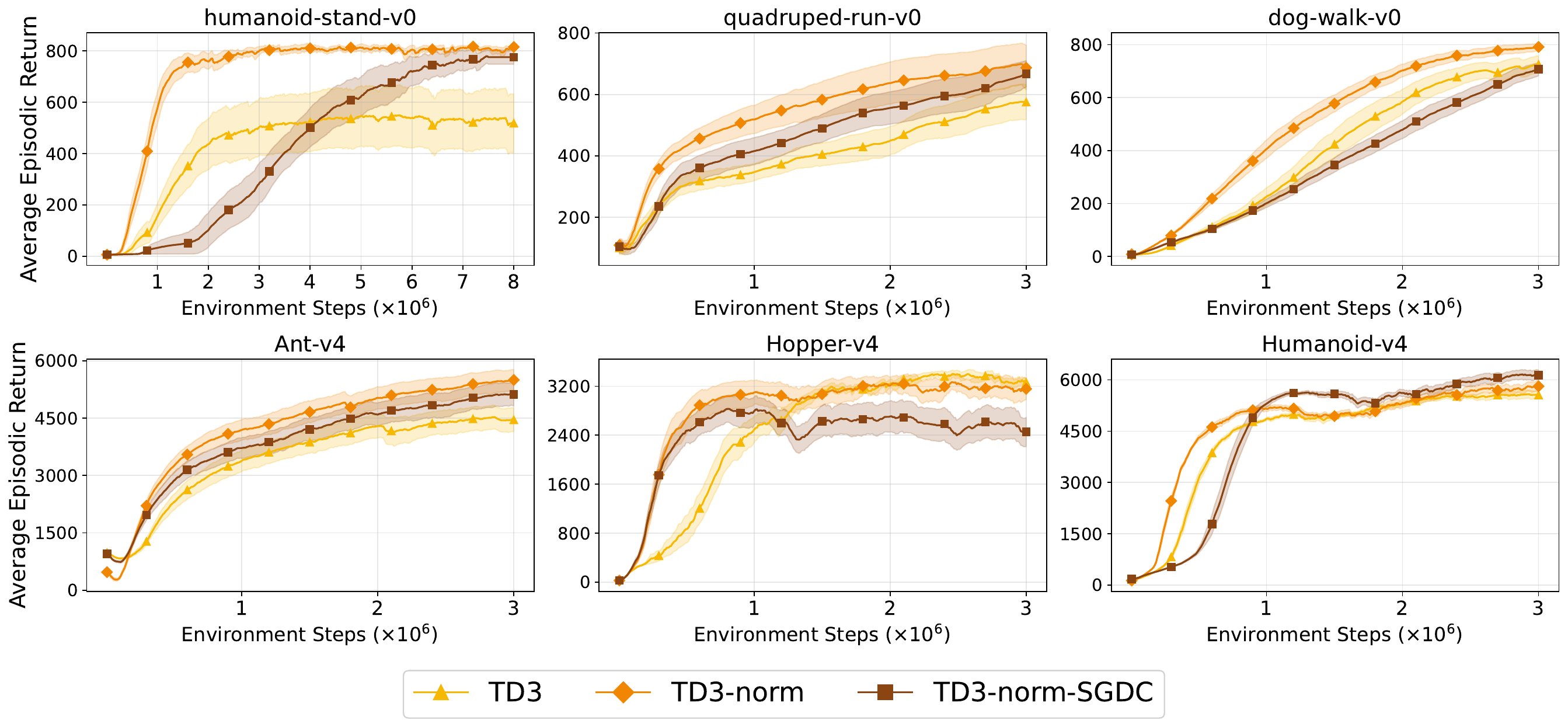}
    \end{center}
    \caption{Batch performance of TD3-norm when the critic is optimized with SGDC rather than
    Adam, on MuJoCo Gym and DM Control Suite tasks. One panel per environment; each curve is
    the mean undiscounted return over 10 seeds and the shaded band a 95\% confidence interval.
    Final performance is matched; the SGDC runs are slower to reach it on
    \texttt{humanoid-stand-v0}, which is shown over 8M steps rather
    than 3M for that reason.}
    \label{fig:batch_supp_mix}
\end{figure}

\autoref{fig:batch_supp_mix} reports TD3-norm pre-trained with SGDC \citep{sun2025revisiting}
against the Adam-optimized version of \autoref{subsec:supp_batch_norm}. Final performance is
matched, while sample efficiency not always: on \texttt{humanoid-stand-v0} the
SGDC runs converge more slowly, and we extend the training horizon from 3M to 8M steps for
this environment so that the agent reaches a comparable level. The same substitution
applied to SAC is reported in \autoref{sec:supp_sac_ft}.

\clearpage
\newpage
\section{Perturbed environments details}
\label{sec:supp_modenv}

Concerning the \texttt{dog-walk-v0} environment of DeepMind Control Suite \citep{tunyasuvunakool2020dm_control}, the following changes have been applied to obtain a perturbed version \texttt{dog-walk-v0-p} of the original environment:

\begin{enumerate}
    \item The control range is changed from $[-1,1]$ to $[-0.7,0.7]$.
    \item The stiffness of the joint \textit{lumbar\_extend} is changed from $30.0$ to $5.0$.
    \item The stiffness of the joint \textit{lumbar\_bend} is changed from $30.0$ to $5.0$.
    \item The stiffness of the joint \textit{cervical} is changed from $4.0$ to $1.0$.
    \item The stiffness of the joint \textit{hip} is changed from $5.0$ to $2.0$.
    \item The control gain for the knee joints is reduced from $30$ to $10$.
    \item The control gain for the ankle joints is reduced from $20$ to $10$.
\end{enumerate}

Collectively, these modifications simulate either an ageing dog or the degradation of materials and actuators in a robotic dog. For instance, reduced lumbar stiffness reflects the effect of weakened back muscles in a biological dog, or material fatigue in the spinal structure of a robotic one.

Concerning \texttt{walker-run-v0} we apply similar modifications to derive two perturbed versions, \texttt{walker-run-v0-p1} and \texttt{walker-run-v0-p2}. Concerning \texttt{walker-run-v0-p1} the modifications are symmetric for the two legs:

\begin{enumerate}
    \item The stiffness of the ankles is changed from $0.0$ to $15.0$.
    \item The stiffness of knees are changed from $0.0$ to $15.0$.
    \item The actuator gains for right and left hips are reduced from $100$ to $80$.
    \item The actuator gains for right and left knees are reduced from $50$ to $40$.
    \item The actuator gains for right and left ankles are reduced from $20$ to $16$.
\end{enumerate}

While for \texttt{walker-run-v0-p2}, an asymmetric perturbation is applied: 

\begin{enumerate}
    \item The stiffness of the ankles is changed from $0.0$ to $15.0$.
    \item The stiffness of knees are changed from $0.0$ to $15.0$.
    \item The actuator gain for the  left hip is reduced from $100$ to $60$.
    \item The actuator gain for the  left knees is reduced from $50$ to $30$.
    \item The actuator gain for the  left ankle is reduced from $20$ to $12$.
\end{enumerate}

\begin{figure}[h]
    \begin{center}
        \includegraphics[width=1.0\textwidth]{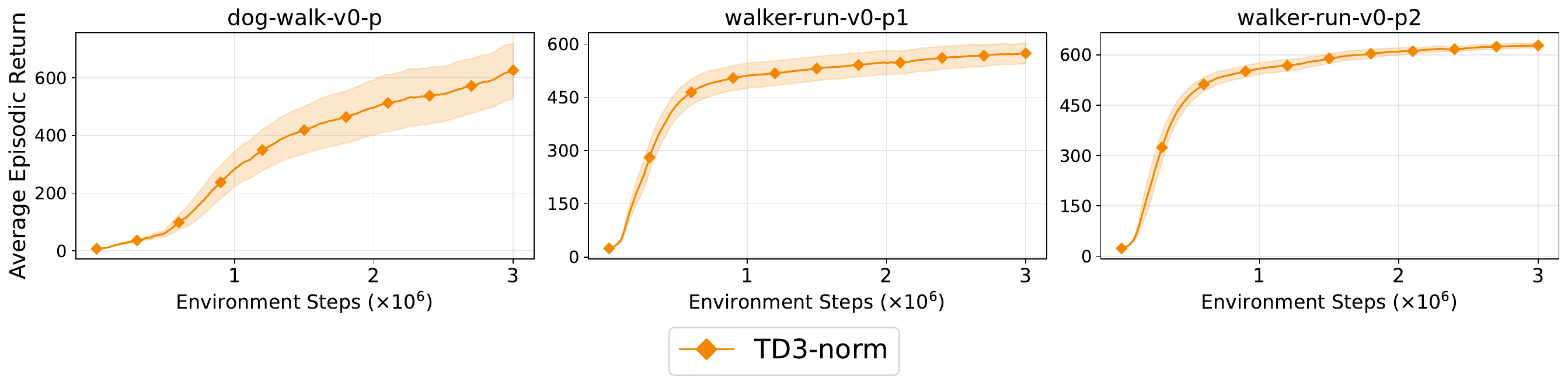}
    \end{center}
    \caption{Results for TD3-norm trained \textit{from scratch} in the perturbed environments.}
    \label{fig:td3_mod_envs}
\end{figure}

\begin{figure}[h]
    \begin{center}
        \includegraphics[width=1.0\textwidth]{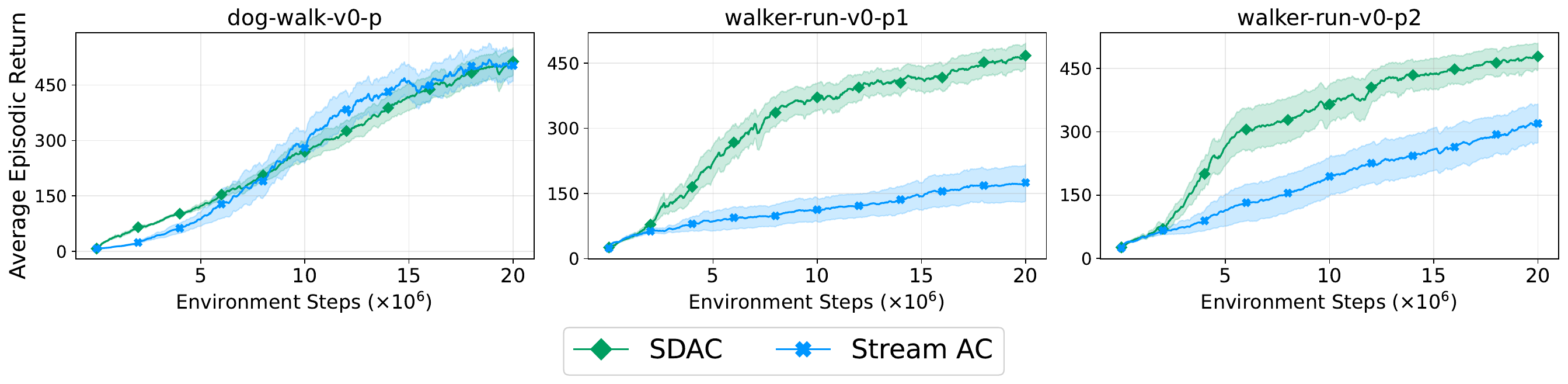}
    \end{center}
    \caption{SDAC and Stream AC($\lambda$) trained \emph{from scratch} directly in the
    perturbed environments, over 20M steps. Mean undiscounted return over 10 seeds, shaded
    bands are 95\% confidence intervals. The final value of SDAC in each panel is
    the level drawn as the grey dash-dotted line in \autoref{fig:ft_main}, which is the
    reference the finetuned agent is compared against.}
    \label{fig:stream_mod_envs}
\end{figure}

For reproducibility reasons, we provide the modified \texttt{.xml} files for both environments in the code repository. For completeness, we report in \autoref{fig:td3_mod_envs} and \autoref{fig:stream_mod_envs} the performance respectively of TD3-norm and SDAC compared to Stream AC($\lambda$) trained from scratch directly in the perturbed environments. These results provide a direct reference for assessing the quality of the finetuning.
\clearpage 

\clearpage
\newpage
\section{Additional finetuning details}
\label{sec:supp_ft}

In this section we report all details necessary to reproduce the finetuning experiments. As specified in \autoref{sec:intro} and \autoref{subsec:finetuneexp}, our experiments consider both the Sim2Real and the batch-to-streaming alternation scenarios. The scenarios differ in which checkpoints are used: for \textit{Sim2Real} we consider early (e.g., 500K steps), intermediate (1.5M steps), and final checkpoints, whereas for batch-to-streaming alternation only early checkpoints are used. The checkpoints used for the finetuning experiments, along with the corresponding number of pre-training steps, are made available in the code repository, so that the experiments can be reproduced without requiring a GPU for pre-training. We observe that in a batch-to-streaming scenario the finetuning performance may be affected by both the pre-training checkpoint and the finetuning seed. To account for this variability, we present the finetuning results as separate curves, each corresponding to a pre-training checkpoint obtained with a distinct seed, and repeat each finetuning experiment three times to report a 95\% confidence interval. The main paper shows three such curves per environment for visual clarity; in \autoref{subsec:supp_ft_add_res} we report three additional curves per environment, each corresponding to a different pre-training checkpoint. Across the two figures, this amounts to six checkpoints and 18 finetuning runs per environment.

In \autoref{subsec:supp_ft_hyper}, we report all hyperparameters used during both pre-training and finetuning, along with a brief description of the proposed critic warm-up phase. In \autoref{subsec:supp_naive_b2s}, we present an additional experiment on the naive batch-to-streaming scenario, where the critic of TD3-norm is trained with Adam instead of SGDC. In \autoref{subsec:supp_ft_abl} we report ablation studies on the use of the critic warm-up phase and the exploration noise for SDAC during finetuning. Finally, \autoref{sec:supp_sac_ft} presents the results for the SAC-to-S2AC transition.

\subsection{TD3-norm and SDAC hyperparameters}
\label{subsec:supp_ft_hyper}
For pre-training with TD3-norm, we use the hyperparameters defined in \autoref{table:TD3_hyper} for the initial tests with the Adam optimizer, and those in \autoref{table:TD3SGDC_hyper} for SGDC.

Regarding SDAC, since in a \textit{Sim2Real} scenario the distribution shift might affect the actual action-value function $Q$, we additionally propose a critic warm-up phase in which the policy is frozen and only the critic is updated. All results in \autoref{subsec:finetuneexp} use the SDAC hyperparameters in \autoref{table:SDAC_hyper_ft}.

\begin{table}[h]
\caption{TD3-norm with SGDC Hyperparameters.}\label{table:TD3SGDC_hyper}
\centering
\begin{tabular}{cll}
\toprule
& Hyperparameter & Value \\
\midrule
\multirow{4}{*}{No streaming params}
& Buffer size & $10^6$\\
&Batch size & $256$\\
&Exploration time steps & $25\times 10^3$\\
&Target smoothing coeff. & $\tau=0.005$\\
\midrule

\multirow{5}{*}{}
& Frequency policy update & $2$\\
& Target Noise & $\mathcal{N}(0,0.2^2)$\\
&Exploration Noise & $\mathcal{N}(0,0.1^2)$\\
\midrule
\multirow{3}{*}{Actor Optimizer} 
& Optimizer        & Adam~\citep{kingma2014adam} \\
& Learning rate    & $3\text{e}-4$ \\
& Betas            & $\beta_1=0.9, \beta_2=0.999$\\
\midrule
\multirow{3}{*}{Critic Optimizer} 
& Optimizer        & SGDC~\citep{sun2025revisiting} \\
& Learning rate    & $0.5$ \\
& Clipping parameter $h$            & $1.0$\\
\bottomrule
\end{tabular}
\end{table} 

\begin{table}[h]
\caption{SDAC Hyperparameters for finetuning.}\label{table:SDAC_hyper_ft}
\centering
\begin{tabular}{cll}
\toprule
& Hyperparameter & Value \\
\midrule
\multirow{3}{*}{}
& Target Noise      & $\mathcal{N}(0,0.1^2)$ \\
& Exploration noise    & $\mathcal{N}(0,0.1^2)$ \\
& Critic warm-up & $5{,}000$ steps \\
\midrule
\multirow{3}{*}{Actor Optimizer} 
& Optimizer        & Adam~\citep{kingma2014adam} \\
& Learning rate    & $3\text{e}-4/256$ \\
& Betas            & $\beta_1=0.9, \beta_2=0.999$\\
\midrule
\multirow{4}{*}{Critic Optimizer} 
& Optimizer        & ObGD~\citep{elsayed2024streaming} \\
& Learning rate    & $1.0$ \\
& $\lambda$    & $0.8$ \\
& $\kappa$    & $2.0$ \\
\bottomrule
\end{tabular}
\end{table} 

\newpage

\subsection{Additional experiment for the naive batch2stream}\label{subsec:supp_naive_b2s}

\begin{figure}[h]
    \begin{center}
        \includegraphics[width=0.6\textwidth]{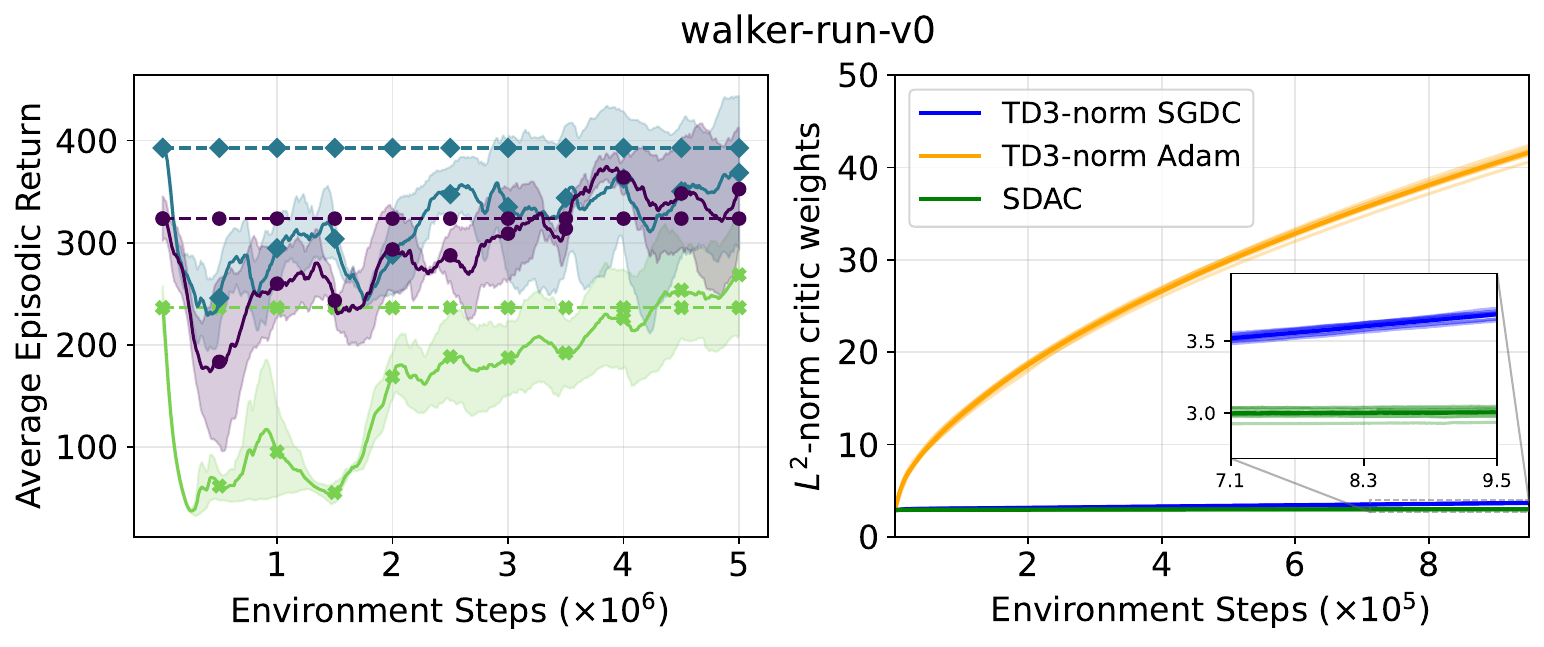}
    \end{center}
    \caption{Left: finetuning performance of SDAC after pre-training with TD3-norm using Adam as critic optimizer. Right: $L^2$-norm of the critic's weights across training.}
    \label{fig:norm_walker}
\end{figure}

Analogously to the discussion in \autoref{sec:b2s}, we report in \autoref{fig:norm_walker} the direct transition results from TD3-norm with Adam to SDAC also in the \texttt{walker-run-v0} environment. Also in this case we can see both the substantial performance drops in the performance, and the huge difference in the $L^2$-norm of the critic weights when using TD3-norm with Adam for pre-training, compared to SGDC, enforcing the analysis carried out in \autoref{sec:b2s}. 

\subsection{Additional pre-training checkpoints}
\label{subsec:supp_ft_add_res}

\begin{figure}[h]
    \begin{center}
        \includegraphics[width=1.0\textwidth]{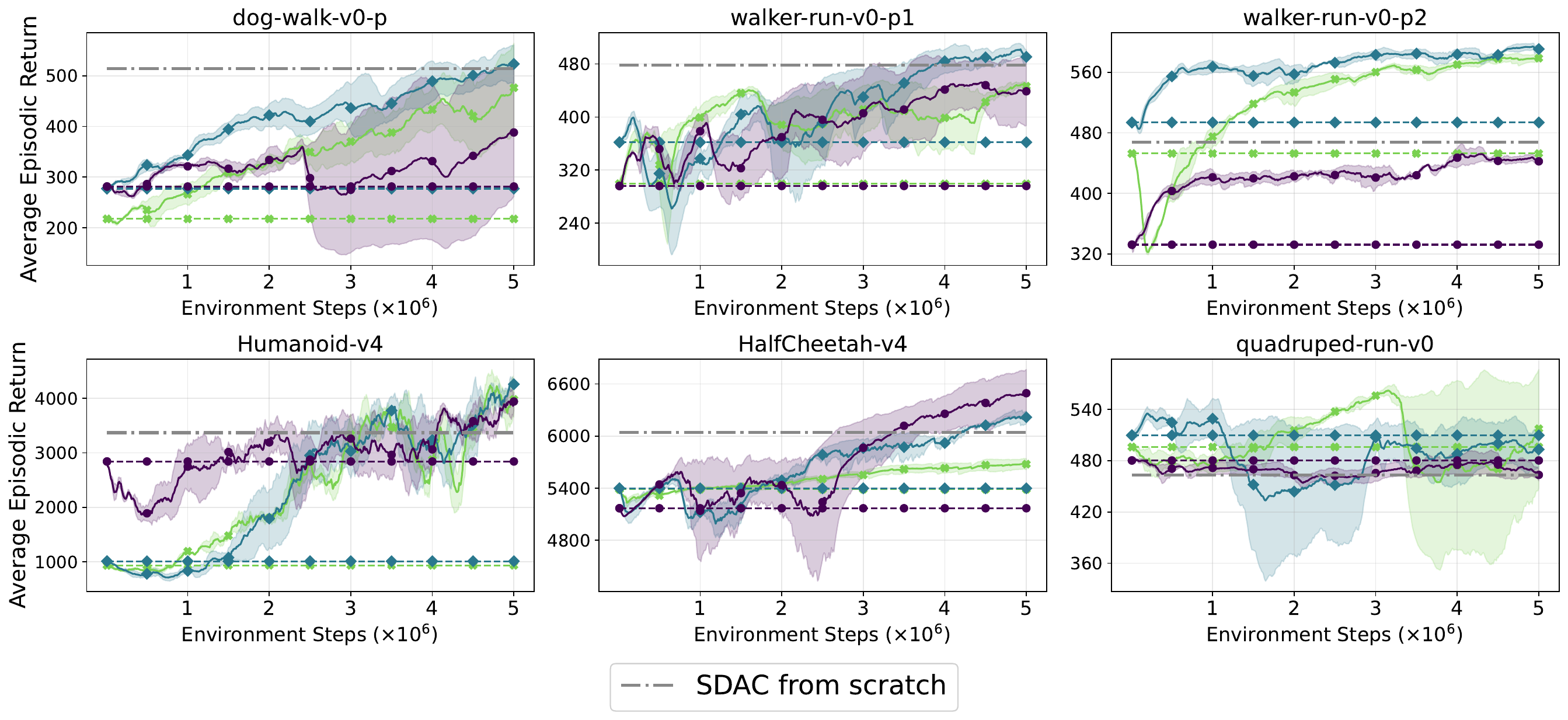}
    \end{center}
    \caption{Three \emph{further} pre-training checkpoints per environment, finetuned with
    SDAC under the same pipeline as \autoref{fig:ft_main}; together the two figures cover six
    checkpoints per environment. Read exactly as \autoref{fig:ft_main}: one coloured curve per
    checkpoint, the dashed line of the same colour is the return that checkpoint had at the
    switch, the grey dash-dotted line is SDAC from scratch after 20M steps, shaded bands are
    95\% confidence intervals over three finetuning seeds, and the horizontal axis counts
    finetuning steps only.}
    \label{fig:finetuning_mixed_supp}
\end{figure}

As mentioned in \autoref{subsec:finetuneexp}, we report in \autoref{fig:finetuning_mixed_supp} three additional curves per environment, each corresponding to a different pre-training checkpoint. As visible from the plots, the results are generally aligned with the ones in \autoref{fig:ft_main}.

\subsection{Finetuning ablations}
\label{subsec:supp_ft_abl}

Finally, we conduct two ablation studies to evaluate the effectiveness of the critic warm-up phase at the beginning of finetuning and the role of exploration noise during finetuning: indeed, while exploration noise is necessary in deterministic approaches, the inherently noisy nature of the online streaming setting may itself act as an implicit exploration mechanism, potentially making explicit noise injection redundant.

\begin{figure}[h]
    \begin{center}
        \includegraphics[width=1.0\textwidth]{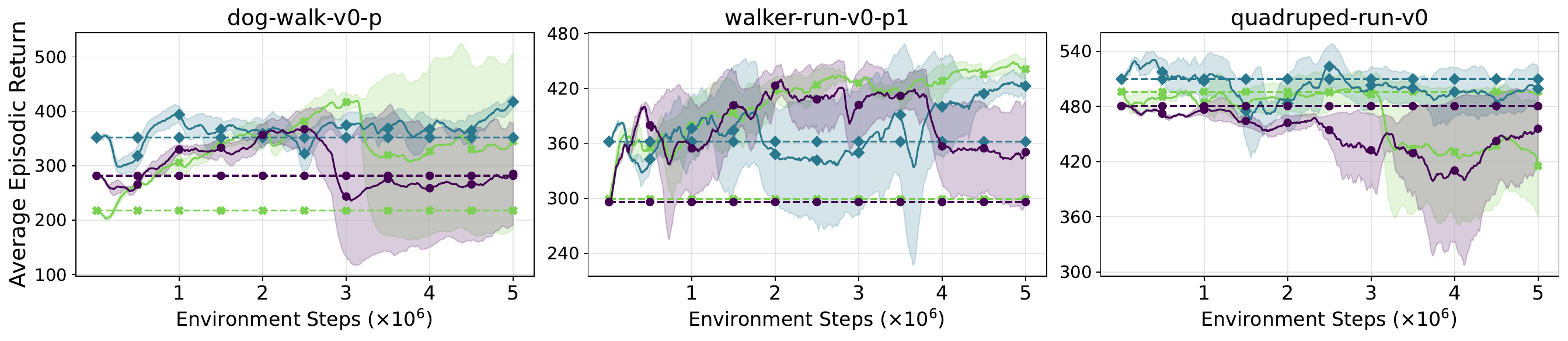}
    \end{center}
    \caption{Ablation of the critic warm-up: the policy
    is unfrozen from the first streaming step instead of after $5{,}000$ critic-only updates.
    Everything else matches \autoref{fig:ft_main}, and the figure is read the same way: one
    coloured curve per pre-training checkpoint, the dashed line of the same colour its return
    at the switch, the grey dash-dotted line SDAC from scratch after 20M steps, shaded bands
    95\% confidence intervals over three finetuning seeds. Comparing a curve here against its
    counterpart in \autoref{fig:ft_main} isolates the contribution of the warm-up.}
    \label{fig:finetuning_nowarmup}
\end{figure}

\begin{figure}[h]
    \begin{center}
        \includegraphics[width=1.0\textwidth]{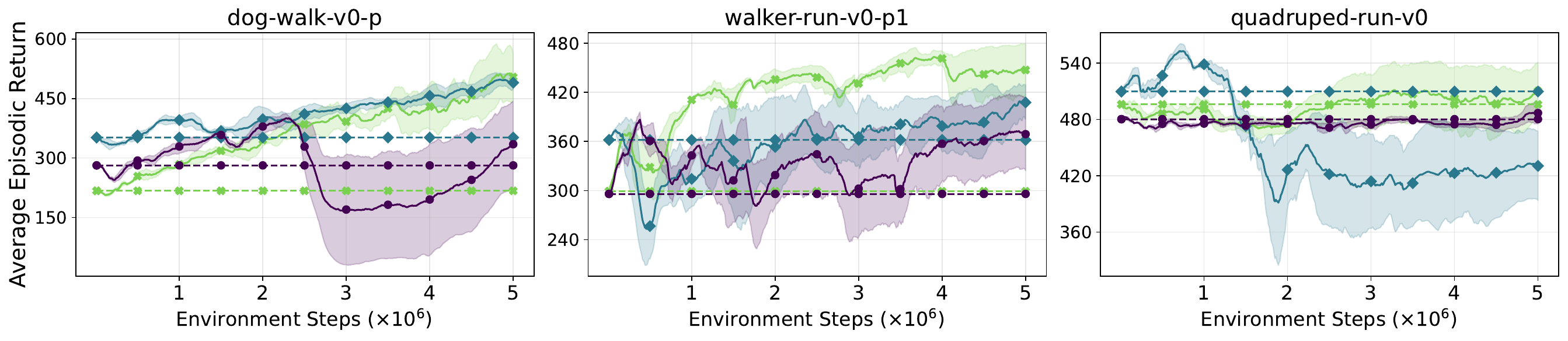}
    \end{center}
    \caption{Ablation of the exploration noise: the deterministic policy is executed without added Gaussian
    noise during finetuning, leaving the stochasticity of single-transition updates as the
    only source of exploration. Read as \autoref{fig:ft_main}, and compare curve by curve
    against it to isolate the contribution of the noise.}
    \label{fig:finetuning_nonoise}
\end{figure}

\autoref{fig:finetuning_nowarmup} and \autoref{fig:finetuning_nonoise} report the two
ablations. Both components contribute to more stable learning during finetuning, which
supports keeping them; neither is responsible for the difference between a surviving and a
collapsing transition, which \autoref{subsec:naive} attributes to the optimizer.

\subsection{SAC-to-S2AC}
\label{sec:supp_sac_ft}

\begin{figure}[h]
    \begin{center}
        \includegraphics[width=0.7\textwidth]{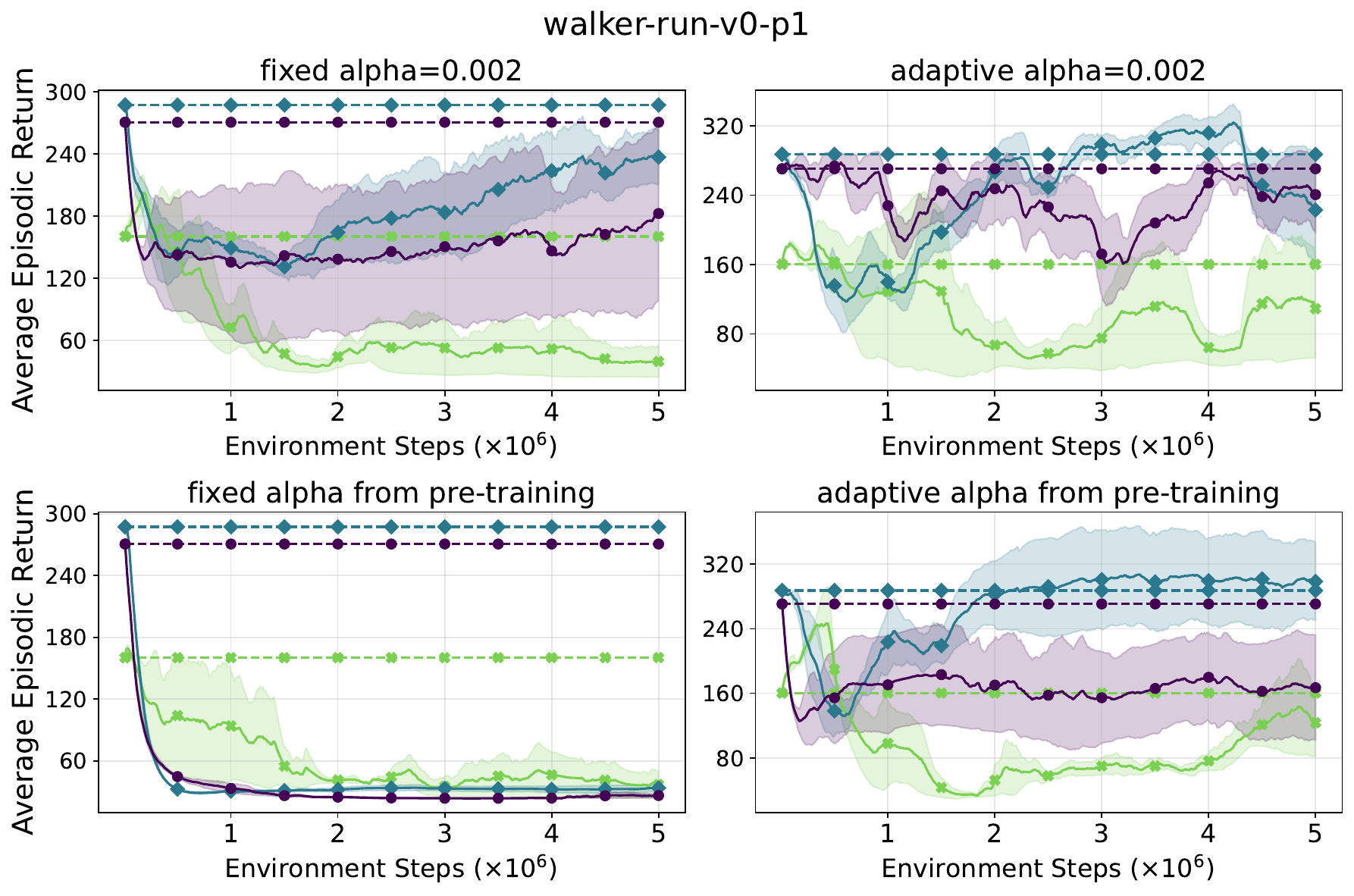}
    \end{center}
    \caption{The maximum-entropy branch of the pipeline on \texttt{walker-run-v0-p1}: S2AC
    finetuned from SAC-norm checkpoints pre-trained with SGDC. The four panels cross the two
    choices left open for the entropy coefficient: columns hold the coefficient fixed or scale it as
    $\alpha \leftarrow \alpha_0/\sigma_r$ during finetuning, rows set the base value $\alpha_0$ to the one
    S2AC uses from scratch ($0.002$) or to the value reached by the autotuner at the end of
    pre-training.
    Within each panel the figure is read as \autoref{fig:ft_main}: one coloured curve per
    pre-training checkpoint, the dashed line of the same colour its return at the switch,
    shaded bands 95\% confidence intervals over three finetuning seeds. No setting keeps all
    three checkpoints at or above their starting return, which is the sense in which this
    branch is not yet reliable.}
    \label{fig:ft_sac}
\end{figure}

\autoref{fig:ft_sac} reports the results of S2AC finetuning in the \texttt{walker-run-v0-p1} environment, where we tested two different values for the base entropy coefficient $\alpha_0$, i.e., the same value used for S2AC from scratch, and the final value of the pre-training stage, where the entropy autotune is employed. Moreover, for both values we tested either keeping them fixed or scaling them dynamically with the same procedure used for S2AC from scratch and detailed in \autoref{sec:method}. As we can see from the plots, the results are not yet reliable, and require further investigation.

\begin{figure}[ht]
    \begin{center}
        \includegraphics[width=0.4\textwidth]{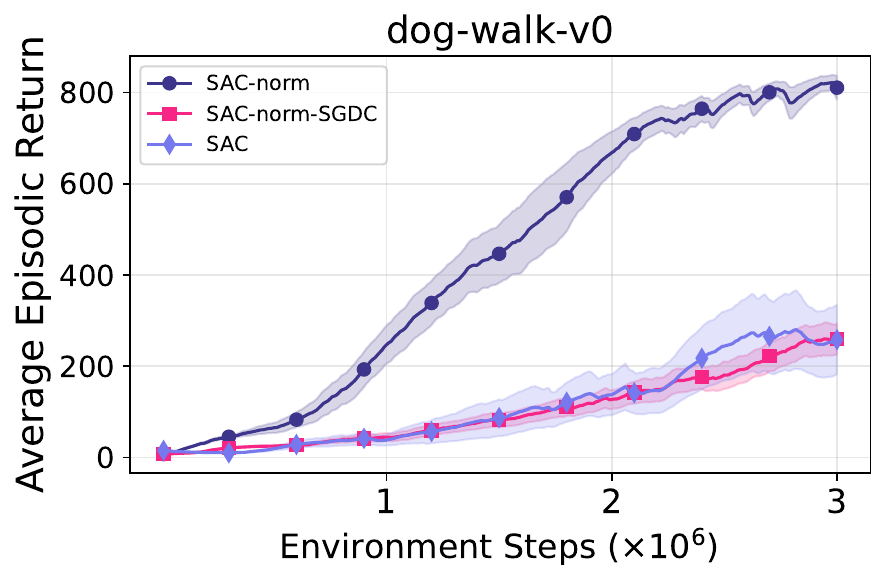}
    \end{center}
    \caption{Batch performance of SAC-norm on \texttt{dog-walk-v0} when the critic is
    optimized with SGDC rather than Adam. Mean undiscounted return over 10 seeds, shaded band a 95\%
    confidence interval.}
    \label{fig:sac_sgdc}
\end{figure}

Finally, to support what discussed in \autoref{subsec:finetuneexp}, pre-training results of SAC with SGDC for the \texttt{dog-walk-v0} environment are reported in \autoref{fig:sac_sgdc}.

%%%%%%%%%%%%%%%%%%%%%%%%%%%%%%%%%%%%%%%%%%%%%%%%%%%%%%%%%%%%

\clearpage
\section*{NeurIPS Paper Checklist}

\begin{enumerate}

\item {\bf Claims}
    \item[] Question: Do the main claims made in the abstract and introduction accurately reflect the paper's contributions and scope?
    \item[] Answer: \answerYes{} % Replace by \answerYes{}, \answerNo{}, or \answerNA{}.
    \item[] Justification:  The claims presented in the abstract and introduction are supported by the experimental results and ablation studies conducted in simulation.

\item {\bf Limitations}
    \item[] Question: Does the paper discuss the limitations of the work performed by the authors?
    \item[] Answer: \answerYes{} % Replace by \answerYes{}, \answerNo{}, or \answerNA{}.
    \item[] Justification: In the conclusions and in the sections presenting the results, limitations are highlighted, bringing attention to the difficulty of transitioning from batch to streaming learning.

\item {\bf Theory assumptions and proofs}
    \item[] Question: For each theoretical result, does the paper provide the full set of assumptions and a complete (and correct) proof?
    \item[] Answer: \answerNA{} % Replace by \answerYes{}, \answerNo{}, or \answerNA{}.
    \item[] Justification: The paper does not include theoretical results.

    \item {\bf Experimental result reproducibility}
    \item[] Question: Does the paper fully disclose all the information needed to reproduce the main experimental results of the paper to the extent that it affects the main claims and/or conclusions of the paper (regardless of whether the code and data are provided or not)?
    \item[] Answer: \answerYes{} % Replace by \answerYes{}, \answerNo{}, or \answerNA{}.
    \item[] Justification: The main paper details the experimental setup and benchmarks, while the appendix covers training details, hyperparameters, hardware specifications, and pseudocode for the proposed methods. The code is available via an anonymous link.

\item {\bf Open access to data and code}
    \item[] Question: Does the paper provide open access to the data and code, with sufficient instructions to faithfully reproduce the main experimental results, as described in supplemental material?
    \item[] Answer: \answerYes{} % Replace by \answerYes{}, \answerNo{}, or \answerNA{}.
    \item[] Justification: The evaluation environments are based on widely-used open-source RL libraries (MuJoCo Gym and DM Control), and the code for the proposed methods is available via an anonymous link.

\item {\bf Experimental setting/details}
    \item[] Question: Does the paper specify all the training and test details (e.g., data splits, hyperparameters, how they were chosen, type of optimizer) necessary to understand the results?
    \item[] Answer: \answerYes{} % Replace by \answerYes{}, \answerNo{}, or \answerNA{}.
    \item[] Justification: We list important hyperparameters, neural network architectures, and other training details in the appendix.

\item {\bf Experiment statistical significance}
    \item[] Question: Does the paper report error bars suitably and correctly defined or other appropriate information about the statistical significance of the experiments?
    \item[] Answer: \answerYes{} % Replace by \answerYes{}, \answerNo{}, or \answerNA{}.
    \item[] Justification: We assume normally distributed errors. All results are averaged over 10 runs and reported with 95\% confidence interval.

\item {\bf Experiments compute resources}
    \item[] Question: For each experiment, does the paper provide sufficient information on the computer resources (type of compute workers, memory, time of execution) needed to reproduce the experiments?
    \item[] Answer: \answerYes{} % Replace by \answerYes{}, \answerNo{}, or \answerNA{}.
    \item[] Justification:  In the appendix, the hardware specifications and usages are specified.
    
\item {\bf Code of ethics}
    \item[] Question: Does the research conducted in the paper conform, in every respect, with the NeurIPS Code of Ethics \url{https://neurips.cc/public/EthicsGuidelines}?
    \item[] Answer: \answerYes{} % Replace by \answerYes{}, \answerNo{}, or \answerNA{}.
    \item[] Justification: Our paper conforms to the NeurIPS Code of Ethics.

\item {\bf Broader impacts}
    \item[] Question: Does the paper discuss both potential positive societal impacts and negative societal impacts of the work performed?
    \item[] Answer: \answerNA{} % Replace by \answerYes{}, \answerNo{}, or \answerNA{}.
    \item[] Justification: This paper proposes a method for finetuning on-device in reinforcement learning, and is not tied to specific applications. As such, it shares the many potential societal consequences that are associated with reinforcement learning and automation as a whole, spanning from environmental impact to concerns on ethics and alignment.
    
\item {\bf Safeguards}
    \item[] Question: Does the paper describe safeguards that have been put in place for responsible release of data or models that have a high risk for misuse (e.g., pre-trained language models, image generators, or scraped datasets)?
    \item[] Answer: \answerNA{} % Replace by \answerYes{}, \answerNo{}, or \answerNA{}.
    \item[] Justification: We do not release high-risk data or models.

\item {\bf Licenses for existing assets}
    \item[] Question: Are the creators or original owners of assets (e.g., code, data, models), used in the paper, properly credited and are the license and terms of use explicitly mentioned and properly respected?
    \item[] Answer: \answerYes{} % Replace by \answerYes{}, \answerNo{}, or \answerNA{}.
    \item[] Justification: We implemented most of our algorithms from scratch and use popular benchmarks and cite them as and when necessary. All our results are generated during training.

\item {\bf New assets}
    \item[] Question: Are new assets introduced in the paper well documented and is the documentation provided alongside the assets?
    \item[] Answer: \answerYes{} % Replace by \answerYes{}, \answerNo{}, or \answerNA{}.
    \item[] Justification: We provide our code and instructions to run the experiments.

\item {\bf Crowdsourcing and research with human subjects}
    \item[] Question: For crowdsourcing experiments and research with human subjects, does the paper include the full text of instructions given to participants and screenshots, if applicable, as well as details about compensation (if any)? 
    \item[] Answer: \answerNA{} % Replace by \answerYes{}, \answerNo{}, or \answerNA{}.
    \item[] Justification: The paper does not involve crowdsourcing nor research with human subjects.

\item {\bf Institutional review board (IRB) approvals or equivalent for research with human subjects}
    \item[] Question: Does the paper describe potential risks incurred by study participants, whether such risks were disclosed to the subjects, and whether Institutional Review Board (IRB) approvals (or an equivalent approval/review based on the requirements of your country or institution) were obtained?
    \item[] Answer: \answerNA{} % Replace by \answerYes{}, \answerNo{}, or \answerNA{}.
    \item[] Justification: The paper does not involve crowdsourcing nor research with human subjects.

\item {\bf Declaration of LLM usage}
    \item[] Question: Does the paper describe the usage of LLMs if it is an important, original, or non-standard component of the core methods in this research? Note that if the LLM is used only for writing, editing, or formatting purposes and does \emph{not} impact the core methodology, scientific rigor, or originality of the research, declaration is not required.
    %this research? 
    \item[] Answer: \answerNA{} % Replace by \answerYes{}, \answerNo{}, or \answerNA{}.
    \item[] Justification: LLMs have been used only for writing, editing, and formatting purposes, and did not impact the core methodology, scientific rigor, or originality of the research.

\end{enumerate}

\end{document}